\pdfoutput=1
\documentclass[final]{neus2025}

\RequirePackage[table]{xcolor} 
\usepackage{tcolorbox} 
\usepackage{soul} 

\usepackage{adjustbox}

\newcommand{\enc}{\operatorname{enc}}
\newcommand{\dec}{\operatorname{dec}}

\title
{Four Principles for Physically Interpretable World Models}
\usepackage{times}

\coltauthor{\Name{Jordan Peper} \thanks{First co-authors: equal contribution.}
\Email{jpeper@ufl.edu}\\
 \Name{Zhenjiang Mao} \footnotemark[1] \Email{z.mao@ufl.edu}\\
 \Name{Yuang Geng} 
  \Email{yuang.geng@ufl.edu}\\
   \Name{Siyuan Pan} 
  \Email{pansiyuan@ufl.edu}\\
 \Name{Ivan Ruchkin} \Email{iruchkin@ece.ufl.edu}\\
 \addr Trustworthy Engineered Autonomy (TEA) Lab, University of Florida, Gainesville, FL, USA}

\begin{document}

\maketitle

\vspace{-6mm}

\begin{abstract}
As autonomous systems are increasingly deployed in open and uncertain settings, there is a growing need for trustworthy world models that can reliably predict future high-dimensional observations.
The learned latent representations in world models lack direct mapping to meaningful physical quantities and dynamics, limiting their utility and interpretability in downstream planning, control, and safety verification. In this paper, we argue for a fundamental shift from \emph{physically informed} to \emph{physically interpretable} world models --- and crystallize \emph{four principles} that leverage symbolic knowledge to achieve these ends: (1) functionally organizing the latent space according to the physical intent, (2) learning aligned invariant and equivariant representations of the physical world, (3)  integrating multiple forms and strengths of supervision into a unified training process, and (4) partitioning generative outputs to support scalability and verifiability. We experimentally demonstrate the value of each principle on two benchmarks. This paper opens several intriguing research directions to achieve and capitalize on full physical interpretability in world models.

\end{abstract}

\begin{keywords}
  world models, representation learning, neuro-symbolic AI, trustworthy autonomy\\
\hspace{0.75mm} \textbf{Source code:} {\small \href{https://github.com/trustworthy-engineered-autonomy-lab/piwm-principles}{https://github.com/Trustworthy-Engineered-Autonomy-Lab/piwm-principles}}
\end{keywords}

\vspace{-4mm}
\section{Introduction}\label{sec:intro}
\vspace{-1mm}

\looseness=-1
Autonomous systems are increasingly deployed in open and uncertain environments~\cite[]{saidi_autonomous_2022,topcu_assured_2020} and use high-dimensional observations to perceive these environments in necessary detail. To achieve high performance, planning and control are often implemented with deep learning methods like reinforcement learning (RL)~\cite[]{yang_safe_2022,garg_comparison_2019}. Since RL training is sample-inefficient, it is impractical to perform in the real world --- leading to controllers trained ``in the imagination'' of \textit{world models}~\cite[]{ha_recurrent_2018,wu_daydreamer_2022}. 

\looseness=-1
World models learn to approximate the physical world by predicting future observations based on current observations and actions. Popular neural world models compress observations into the latent space using an autoencoder, propagate these latent values forward in time based on learned temporal dependencies~\cite[]{deng_facing_2023}, and decode them into predicted observations. World models can be improved by injecting symbolic physical knowledge into their structure and training process. 
For example, \cite{chen_automated_2022} automatically extracted physically meaningful variables from raw observations, yielding more stable long-horizon predictions than standard autoencoders.
\cite{brunton_discovering_2016} similarly used sparse regression to recover governing equations of nonlinear dynamics from noisy data.
Controllers also generalize better when expressed as neuro-symbolic predicates that combine vision-language models with predefined control primitives~\cite[]{liang_visualpredicator_2024}.

\looseness=-1
A major challenge of modern world models is their lack of \textit{physical interpretability}. We define it as the degree to which a model's learned latent space corresponds meaningfully to the underlying physics: (a) how well latent embeddings map to physical variables, and (b) how closely latent dynamics emulate physical processes. Without sufficient physical interpretability, a world model offers limited utility in classical model-based autonomy and the design of physically grounded rewards for RL. We also cannot obtain physical guarantees from reachability analysis based on world models~\cite[]{katz_verification_2022}. The core reason for this uninterpretability is that deep learning thrives on distributed representations, in which each feature is partially encoded in multiple latent variables~\cite[]{hinton_learning_1986}. This challenge is further complicated by partial online observability of the physical state and the difficulty of precisely labeling the data (e.g., indicating which state is riskier in a video).

\looseness=-1
This paper calls for a paradigm shift from 
\textit{physically informed} world models to  \textit{physically interpretable} ones. 
The former use symbolic physical knowledge to make learning more effective, efficient, and generalizable. The latter creates neuro-symbolic latent representations with explicit physical meaning, thus subsuming physically informed approaches. Physically meaningful representations bring in a plethora of desirable qualities such as reliability, verifiability, and debuggability.

By carefully analyzing the existing world model literature, this paper advances \textit{four guiding principles} that underlie physical interpretability of learned world representations. Specifically, each principle asserts that \textbf{physically interpretable world models should:}
\begin{itemize}
    \item \textbf{Principle 1:} \dots have a \emph{functionally organized} latent space.
    \item \textbf{Principle 2:} \dots learn \emph{aligned} invariant and equivariant representations of the physical world.
    \item \textbf{Principle 3:} \dots integrate \emph{multiple forms and strengths} of supervision into training.
    \item \textbf{Principle 4:} \dots \textit{partition} their generative outputs to support scalability and verifiability. 
\end{itemize}

The next section identifies the interpretability gaps in the existing world models, while Section~\ref{sec:principles} details the four principles. In Section~\ref{sec:experiments}, we perform lightweight validation to demonstrate the value of these principles. 
Finally, Section~\ref{sec:future} discusses the newly opened directions for future research.  

\vspace{-2mm}
\section{World Models: State of the Art} \label{sec:relwork}

\looseness=-1
\noindent
\textbf{Foundations of world models.} Modern world models have led to state-of-the-art performance in autonomous planning and control while addressing the data-efficiency concerns of standard RL~\cite[]{deng_facing_2023,micheli_transformers_2023,robine_transformer-based_2023}. Early world models combined a variational autoencoder (VAE) with a recurrent neural network to predict latent dynamics~\cite[]{ha_recurrent_2018}. Later work refined both the encoder-decoder architecture and the surrogate dynamics: PlaNet introduced a recurrent state-space model (RSSM) for prediction~\cite[]{hafner_learning_2019}; Dreamer backpropagated gradients through imagined trajectories to improve latent prediction~\cite[]{hafner_dream_2020}; and DreamerV2 extended the RSSM to categorical latent variables~\cite[]{hafner2022mastering}. More recent research combines autoregressive transformers with self-attention layers to capture detailed temporal dependencies~\cite[]{robine_transformer-based_2023}, or diffusion models to mitigate compounding errors~\cite[]{ding2024diffusion}. World models have been used to optimize planning algorithms for autonomous vehicles in realistic environments: DriveDreamer~\cite[]{wang_drivedreamer_2023} generates realistic video trajectories from multi-modal inputs for policy optimization; DriveWorld~\cite[]{min_driveworld_2024}, OccWorld~\cite[]{zheng_occworld_2023}, UniWorld~\cite[]{min_uniworld_2023}, and RenderWorld~\cite[]{yan_renderworld_2024} forecast detailed 3D occupancy for motion planning.

\begin{figure}[t]
	\centering         \includegraphics[width=\textwidth]{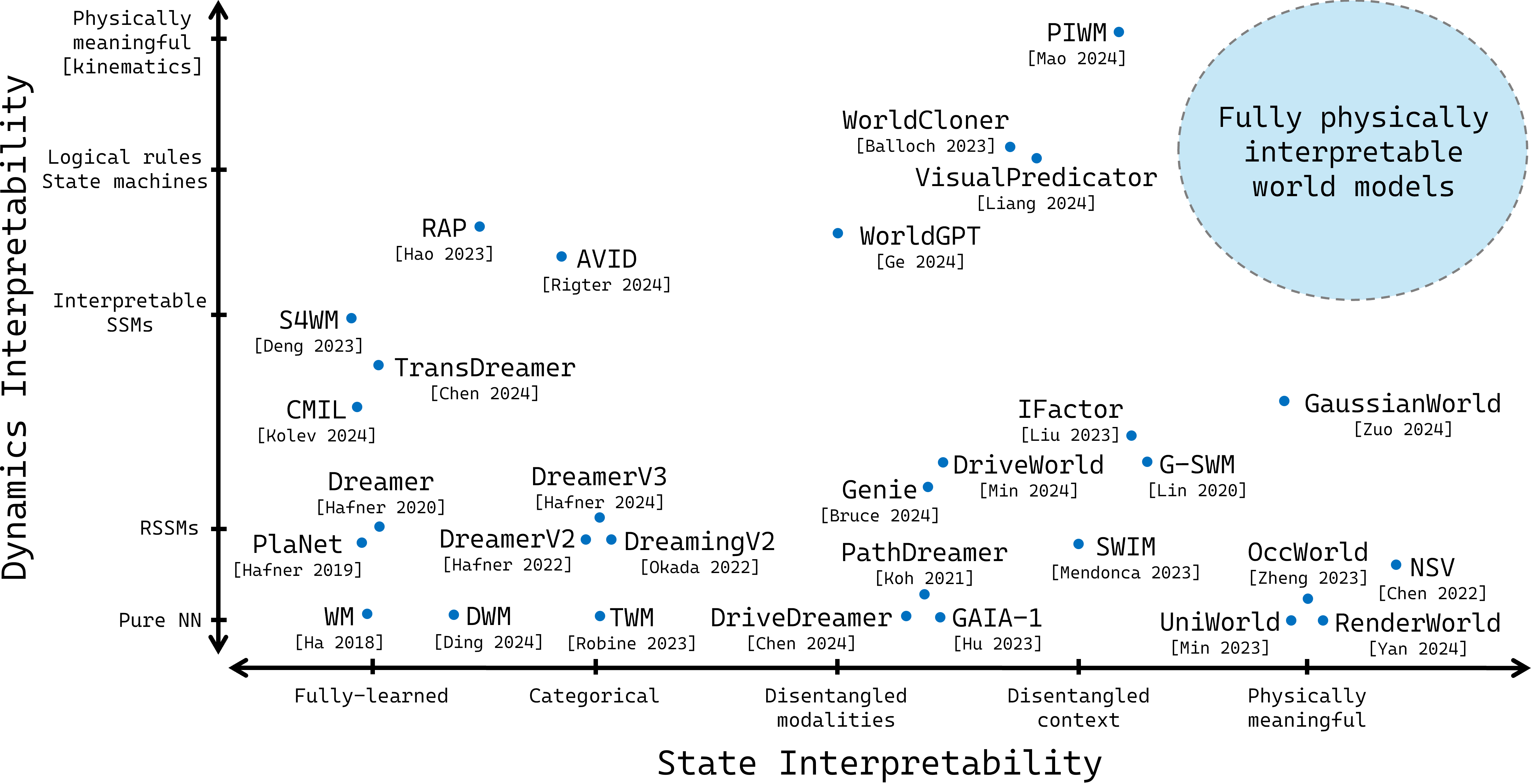}
    \vspace{-8mm}
	\caption{Existing world models by the interpretability of their state and dynamics.}
	\label{fig:wm-interp}
    \vspace{-4mm}
 \end{figure}

\smallskip
\noindent
\textbf{Towards interpretable world models.} Despite the strong performance of world models, their interpretability remains a challenge in most frameworks. Relevant efforts toward disentangling latent variables (i.e., reducing their mutual dependency) include $\beta$-VAEs~\cite[]{higgins_beta-vae_2016} and causal VAEs~\cite[]{yang_causalvae_2021}. This disentanglement strategy is also employed in driving prediction frameworks like GNeVA~\cite[]{lu_towards_2024} and ISAP~\cite[]{itkina_interpretable_2022}. Under the umbrella of world models, G-SWM~\cite[]{lin_improving_2020} investigated a principled modeling framework that inherits interpretable object and context latent separation from various spatial attention approaches~\cite[]{kosiorek_sequential_2018, kossen_structured_2020, jiang_scalor_2020, crawford_exploiting_2020}. \cite{fremont_scenic_2019} proposed \texttt{SCENIC} --- a probabilistic program for generating realistic scenes with physical constraints.
More recent methods impose physical constraints for system identification~\cite{sridhar_guaranteed_2023}, motion prediction~\cite{tumu_physics_2023}, and learnable ODE modeling~\cite{linial_generative_2021, zhong_pi-vae_2023, mao_phy-taylor_2025}.
Incorporating partial knowledge of physics with weak supervision has also improved both the state and dynamics interpretability~\cite[]{mao_towards_2024}. A recent Nature article leveraged the biological alignment of latent representations to predict microbiome community interactions and antibiotic resistance~\cite[]{baig_autoencoder_2023}. \textit{Neuro-symbolic world models} have also begun to emerge: VisualPredicator~\cite[]{liang_visualpredicator_2024} learns a set of abstract states and high-level actions for strong out-of-distribution generalization, whereas WorldCloner~\cite[]{balloch_neuro-symbolic_2023} learns symbolic rules to adapt the dynamics to open world novelty. Relevant neuro-symbolic research includes PhysORD~\cite[]{zhao_physord_2024}, which embeds physical laws into neural models, and work by~\cite{miao_dashcam_2025} transforming dashcam footage from a driving environment into a \texttt{SCENIC} script through a vision-language model.

\looseness=-1
\noindent
\textbf{Knowledge gap.} We observe the lack of world models with full physical interpretability, as per Figure~\ref{fig:wm-interp} (the underlying literature is listed in Table~\ref{tab:review} in the Appendix). Some existing neuro-symbolic architectures scrape the threshold of physically interpretable dynamics, yet lack fluid state representations. On the other hand, multimodal transformer-based architectures preserve the physical context through 3D occupancy but predict with black-box mechanisms. Bridging this gap is key to transitioning from merely \textit{physically informed} world models to fully \textit{physically interpretable} ones. Our recent work highlights the need to address these open questions~\cite[]{lu_surveying_2024} with predictive world models~\cite[]{mao_how_2024} and their foundation-model variants~\cite[]{mao_zero-shot_2024}. 

\smallskip
\noindent
\textbf{Benefits of Physical Interpretability.}
Aligning world models with fundamental physical principles (e.g., kinematics and conservation laws) has been shown to improve their out-of-distribution generalization and robustness~\cite[]{mao_towards_2024, lin_improving_2020, liang_visualpredicator_2024, balloch_neuro-symbolic_2023, greydanus_hamiltonian_2019}. These principles prevent latching onto spurious correlations in training and constrain the models to traverse a physically meaningful manifold when extrapolating observations.
Going further, physically interpretable representations would lead us to a \textit{qualitatively new level} of safety and trustworthiness. It would make world models more transparent and debuggable by cross-checking them with real-world physics. It would also make generative components suitable for closed-loop verification of physical properties. Finally, physical representation would improve RL sample efficiency by shrinking the search space to physically feasible solutions. 

\vspace{-2mm}
\section{Physical Interpretability Principles for World Models}\label{sec:principles}

\begin{figure}[t]
	\centering         \includegraphics[width=\textwidth]{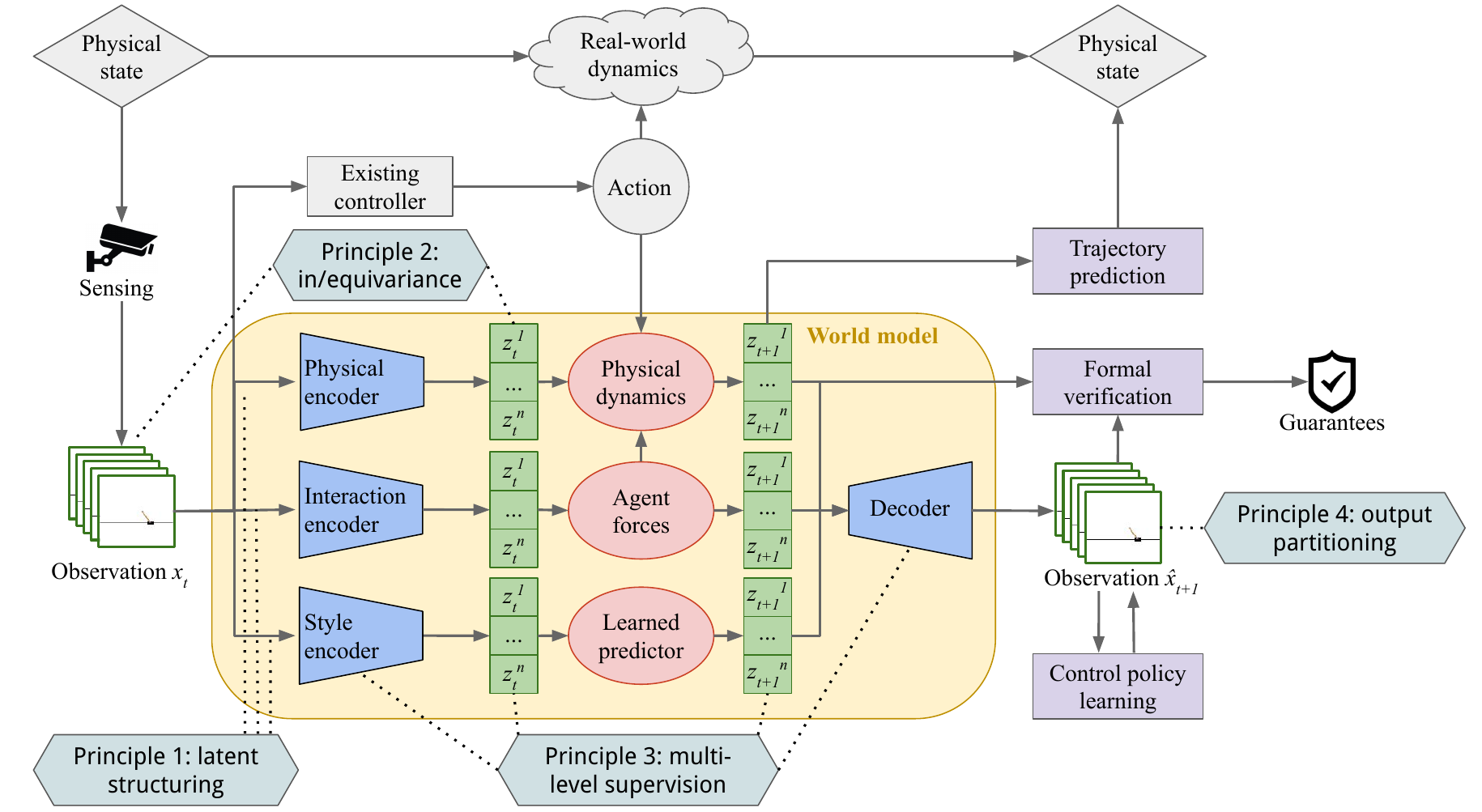}
	\caption{Overview of physically interpretable world models and four principles. 
    }
	\label{fig:overview}
    \vspace{-2mm}
 \end{figure}

This section puts forward \emph{four guiding principles} for building physically interpretable world models. We begin by formally defining a world model:
 
\begin{definition}[World Model]
\label{def:wm}
A \emph{world model} is a function $f : X \rightarrow X$ that maps an observation $x_t \in X \subset \mathbb{R}^n$ to $x_{t+1} = f(x_t) = (\dec \circ \operatorname{dyn} \circ \enc)(x_t)$, where $t$ is the discrete time index, $\enc : X \rightarrow Z$ maps the observation to a latent embedding $z_t \in Z \subset \mathbb{R}^m$, $\operatorname{dyn}$ is the latent dynamics propagating $z_t$ to $z_{t+1}$, and $\dec : Z \rightarrow X$ maps  embedding $z_{t+1}$ to  observation $x_{t+1}$. 

\end{definition}

\looseness=-1
A world model learns its latent space by minimizing the gap between predicted and actual observations. However, due to being a black box, its latent space often lacks direct physical interpretability. To address this, we introduce a general concept of a physically interpretable world model:

\begin{definition}[Physically Interpretable World Model]
\label{def:piwm}
A world model $f$ is \emph{physically interpretable} if: (i) there exists a latent-to-physical mapping $v : \mathbb{R}^m \rightarrow \mathbb{R}^k$, where $k$ is the degrees of freedom of the physical environment, such that $z \xrightarrow{v} z_{phys}$, where $z_{phys} \in \mathbb{R}^k$ is the minimal set of state variables required to fully describe the environment's true dynamics $\operatorname{dyn}_{phys} : \mathbb{R}^k \rightarrow \mathbb{R}^k$; and (ii) it holds that: $v(\operatorname{dyn}(z)) = \operatorname{dyn}_{phys}(v(z))$ for all $z \in Z \subset \mathbb{R}^m$.
\end{definition}



\subsection{Principle 1: Functionally Organizing the Latent Space with Prior Knowledge}


We propose \emph{functionally organizing} a world model by modularizing the latent space and processing embeddings through separate branches, as seen in Figure~\ref{fig:overview}. Each latent state is a vector $z$, which contains $n$ distinct representations of a single observation, each dedicated to unique functionality. Let $x$ represent the world model inputs (e.g., images), and $\enc_i(x)=z_i$ represent the encoder for a particular latent branch $f_i, i = 1..n$, as in Figure~\ref{fig:overview}. Thus, the structured latent space becomes:
\begin{align*}
    z = [\enc_1(x) \quad \enc_2(x) \quad \dots \quad \enc_n(x) ]  
\end{align*}

\noindent
\textbf{Example.} An autonomous driving engineer designs a world model with three branches: (1) absolute dynamics of the agent and environment itself, (2) relative dynamics between other agents, and (3) residual yet relevant features of the surroundings. Let $L$ denote a loss function over $f(z)$ and $x$. The overall training loss should be proportional to the losses in each workflow branch:
\begin{align*} 
    \mathcal{L} \propto L_1(f_1(\enc_1(x)),x) + L_2(f_2(\enc_2(x)),x) + L_3(f_3(\enc_3(x)),x)
\end{align*}

\looseness=-1
Recent work for the first branch aligned latent representations with physical properties~\cite[]{mao_towards_2024}. For the second branch, earlier studies demonstrated that physical interactions between agents can be learned without supervision through graph neural networks (GNNs)~\cite[]{kipf_neural_2018}. In the context of world models, \cite{lin_improving_2020} constructed a separate latent representation using a GNN to capture agent occlusions and interactions. Physics-informed neural networks~\cite[]{raissi_physics-informed_2019, saemundsson_variational_2020} can also improve the physical interpretability of the world model's dynamics. For instance, Hamiltonian neural networks~\cite[]{greydanus_hamiltonian_2019} learn and adhere to physical conservation laws, leading to impressive generalization.  The third branch follows the typical strategy for creating uninterpretable world models and is considered a useful layer to the structured latent space~\cite[]{lin_improving_2020}.
Latent space structuring has become increasingly prevalent to improve the performance of planning and control. For instance, a goal-based neural variational agent (GNeVA) uses separate polyline embeddings for the agent and the map, enabling interpretable generative motion prediction~\cite[]{lu_towards_2024}. Similarly, an interpretable car trajectory prediction framework was proposed, integrating three distinct workflow branches: agent states, high-definition maps, and social context~\cite[]{itkina_interpretable_2022}.

\begin{tcolorbox}[colback=blue!10]
\centering \ul{Principle 1:} Physically interpretable world models should have a \emph{functionally organized} latent space.
\end{tcolorbox}

\subsection{Principle 2: Exploiting Invariances and Equivariances in Input and Latent Spaces}

Neural networks' impressive performance is due in part to their ability to learn rich \emph{distributed representations} from training data. Rather than memorizing examples, these models construct hierarchical feature embeddings that capture data patterns and generalize to i.i.d. samples~\cite[]{hinton_learning_1986}. Nevertheless, training a model to internalize and imagine the world in a human-like manner far from trivial~\cite[]{ha_recurrent_2018}. Encoding high-dimensional observations (e.g., images) through commonplace embedding methods (e.g., through autoencoders or encoder-only transformers) leaves the latent representation generally uninterpretable and task-agnostic. This raises concerns about whether spurious correlations distort the latent space or if it effectively encodes the details necessary for discriminating between features that should remain \emph{functionally disentangled}.

\textit{Invariance and equivariance relations} can help address uninterpretability in representation learning. These terms characterize how representations respond to observation-space transformations. If the representation of $x$ shifts in an expected manner due to a transformation $g(x)$, then the representation model is said to be \textit{equivariant} to that transformation. Likewise, if the representation does not shift under the transformation, then the model is said to be \textit{invariant} to the transformation. For example, bisimulation metrics help learn latent obstacle representations invariant to changes in type, size, and brightness~\cite[]{zhang_invariant_2020}.
\cite{pol_plannable_2020} use contrastive loss to enforce action equivariance. Yet, integrating expert priors remains difficult; one method maps complex observation transformations to simpler latent ones via a symmetric embedding network~\cite[]{park_learning_2022}.

We categorize representations along two dimensions: (1) the nature of their transformation response (invariance versus equivariance) and (2) their degree of human alignment (aligned versus misaligned). A representation that is \emph{aligned-invariant} remains unchanged when an observation undergoes a meaning-preserving transformation, while an \emph{aligned-equivariant} representation transforms predictably when the observation's meaning is altered. In contrast, a representation is \emph{misaligned-invariant} if it does not change under meaningful effects made to the observation (suggesting underfitting), and it is \emph{misaligned-equivariant} if it changes in response to an observation transformation that should not affect the underlying meaning (suggesting a domain shift). Our training objective is to achieve invariance and equivariance alignment by ensuring that the post-transformation representations accurately reflect our human interpretation of the change.

\begin{tcolorbox}[colback=blue!10]
\centering \ul{Principle 2:} Physically interpretable world models should learn \textit{aligned invariant} and \textit{aligned equivariant} representations of their environment.
\end{tcolorbox}

\begin{definition}[Equivariance]\label{def:eqv}
Let $g_\Theta : X \rightarrow X$ be an observation space transformation randomly parameterized by $\Theta$, and let $h_\Phi : Z \rightarrow Z$ be a latent space transformation randomly parameterized by $\Phi$. An encoder $\enc : X \rightarrow Z$ is an \emph{equivariant} function if $\enc(g_\Theta(x)) \overset{d}{=} h_\Phi(\enc(x))$. Invariance is a special case of equivariance where $h_\Phi$ returns its argument.
\end{definition}

Following Definition~\ref{def:eqv}, a simple loss function promotes aligned invariance and equivariance:
$$ \mathcal{L}_{wm}(x) \propto \mathbb{E}_{\Theta, \Phi} \left[ \| \enc(g_{\Theta}(x)) - h_{\Phi}(\enc(x)) \|^2_{2} \right],$$
where $\mathcal{L}_{wm}$ is the overall WM training loss, which is a function of the input observation $x$.


\smallskip
\noindent
\textbf{Example.} 
 Consider a vision-based autonomous car that hands over its neural-based controls to a simpler safety controller if a collision is predicted by its world model, consisting of a ``physical" and ``style" branch per Principle 1. Based on the prior knowledge, an engineer decides that scene brightness should not affect the physical latents; hence, the physics encodings should be \textit{invariant} to changes in observation brightness. However, the style encodings should represent brightness in the resulting latent embedding and, thus, should be equivariant to the changes in brightness.



\subsection{Principle 3: Multi-Level and Multi-Strength Supervision for Latent Representations}
To bridge rich observations and physical meaning, world models must adapt to supervision signals of varying form and strength~\cite[]{lee2013pseudo,chen2020simple} --- from exact state labels to trajectory-level constraints and weak self-supervision.
They must also integrate these signals based on abstraction level (e.g., exact values, intervals, or missing data) to align representations with physical systems.
Multi-level supervision tailors the loss functions and training process to the \textit{level of abstraction} (e.g., full trajectories vs. specific state dimensions) and \textit{strength} (e.g., exact labels vs. intervals). For instance, physical state labels allow for the direct alignment of latent representations with real-world quantities using supervised loss. When such labels are unavailable, temporal consistency and smoothness of trajectories can serve as implicit regularization techniques to constrain learned representations. Finally, self-supervision can leverage data-driven structures to discover meaningful latent representations in entirely unsupervised settings.

\begin{tcolorbox}[colback=blue!10]
\centering \ul{Principle 3:} Physically interpretable world models should integrate multiple \textit{forms} and \textit{strengths} of supervision based on their availability and informativeness.
\end{tcolorbox}

\noindent
\textbf{Supervised Learning:} Strong supervision directly aligns specific latent dimensions with known physical states (e.g., positions, velocities), enabling fine-grained interpretability. In many cases, supervision signals are introduced directly into the embeddings to capture key features from labeled data~\cite[]{zhuang2015supervised}. For example, in low-dimensional systems with position and velocity states \(s = [p, v]\), additional latent dimensions (\(z_{\text{extra}} \sim \mathcal{N}(0,1)\)) can improve reconstruction quality and stability~\cite[]{chen2016infogan,alemi2018fixing,rezende2014stochastic}.

\looseness=-1
\noindent
\textbf{Semi-Supervised Learning:} When the labels are only available for some data, semi-supervised techniques can refine representations. Pseudo-labeling (e.g., Mean Teacher~\cite[]{tarvainen2017mean} and FixMatch~\cite[]{sohn2020fixmatch}) utilizes both labeled and unlabeled data to iteratively improve the latent space. In Motion2Vec~\cite[]{tanwani2020motion2vec}, a small amount of labeled data is first used to initialize the embedding space; subsequently, RNNs predict pseudo-labels for unlabeled data, allowing the model to iteratively refine both the embedding and segmentation components.

\looseness=-1
\noindent
\textbf{Weak Supervision:} Noisy or coarse labels, such as position constraints \((p \in [a, b])\), can be utilized via the trajectory smoothness loss: $\mathcal{L}_{\text{smooth}} = \sum_{t} \| p_t - 2p_{t+1} + p_{t+2} \|^2$. 
Temporal models like Kalman filters~\cite[]{kalman1960new} stabilize noisy trajectories in tasks such as autonomous driving. Interval signals as weak supervision can be directly incorporated into the loss~\cite[]{mao_towards_2024} or combined with contrastive learning to reinforce constraints~\cite[]{sorokin2017end}.

\noindent
\textbf{Self-Supervised Learning:} In the absence of labels, contrastive learning~\cite[]{chen2020simple} aligns latent representations with task-specific similarity metrics (e.g., Euclidean distance or structural similarity).  Contrastive world models~\cite[]{poudel2022contrastive} explicitly employ representation learning losses to map similar states closer in the latent space. Plan2Explore~\cite[]{sekar2020planning} generates self-supervised uncertainty-driven objectives to guide the representations.

\noindent
\textbf{Combining Supervision Levels}: 
For a given dataset \(\mathcal{D}\) with supervision signals (e.g., full trajectories, state variables, or interval constraints), the training objective should integrate matching losses to align the latent space with physical semantics.
We advocate for using every available supervision.
Given explicit state labels, use direct supervision. When only partial information is available, use weakly supervised constraints to refine the representations. 


\subsection{Principle 4: Output Space Partitioning for Verifiability}

\looseness=-1
Ensuring the safety of vision-based autonomy is a critical and open challenge~\cite[]{vision_trajectoy,geng2024bridging,fremont_scenic_2019}. 
The verification of such systems is difficult due to high-dimensional image inputs: classical techniques cannot handle this complexity, motivating new principles for pre-deployment safety guarantees~\cite[]{Althoff_verify, closed_loop_verify}. One such attempt is to employ a generative image model to overapproximate observations from a given physical state and feed them into a state estimator or controller~\cite[]{katz_verification_2022,chuchu_image_controller}. Sadly,  
due to uninterpretable latent states, such ``verification modulo generative models'' does not provide guarantees for the physical world. Furthermore, decoder verification does not scale to large images. 

\looseness=-1
To reduce decoder complexity, we propose to \textit{partition the generated image} into physically meaningful parts. Specifically, a world model will contain multiple generators of output signals --- each dedicated to its own object in the image. Each generator would be separately verifiable, and the results would be combined to provide world model-wide guarantees. This principle reduces each generator's size, making the verification of such world models tractable.

When applied to physically interpretable latent states, this principle can transfer verification guarantees to the physical world: the generators would represent the relationship between images and physical states, not uninterpretable latent ones. Specifications are written at the level of interpretable states (rather than images), after the information has been propagated through perception, control, and dynamics. This process grounds the verification in physical states rather than images.

\begin{tcolorbox}[colback=blue!10]
\centering \ul{Principle 4:} Interpretable world models should \textit{partition generated observations} into segments from multiple simpler generators, enabling scalable verification.
\end{tcolorbox}

\begin{definition}[Partitioned World Model Generation]\label{def:world_model_generation}
A world model decoder $\dec$ translates a latent state $z$ into a generated high-dimensional observation $\hat{x}$, expressed as $\dec(z) = \hat{x}$, by minimizing the reconstruction error between the original and reconstructed observations. Each image segment is produced by a separate decoder: $\dec_1(z) = \hat{x}_1, \dec_2(z) = \hat{x}_2, ..., \dec_n(z) = \hat{x}_n$. The combined generated image is represented as  $\hat{x} = \bigoplus_{i=1}^{n} \hat{x}_i$, where \( \bigoplus \) is a signal composition operation (e.g., overlaying image segments). The corresponding loss function $\mathcal{L}_{\text{gen}}$ is:
\vspace{-2mm}
\begin{equation*} \label{verifiable_loss}
    \mathcal{L}_{\text{gen}} =||x-\hat{x}||^2 +  \lambda \sum_{i=1}^{N} || x_i - \hat{x}_i ||^2 
\end{equation*}
\end{definition}
\vspace{-2mm}

The question of automatic partitioning of world model outputs can be answered by zero-shot approaches like the Segment Anything Model (SAM)~\cite[]{SAE}.
Recently, SAM was used to segment images to improve image and safety prediction~\cite[]{mao_zero-shot_2024}.
A similar partitioning was used in the action space to scale up the verification of vision-based controllers via multiple low-dimensional approximations~\cite[]{geng2024bridging}.
Principle 4 propagates the physical meaning from different parts of the world model (established in Principle 1) to its generative outputs, effectively linking the high-dimensional observation with a lower-dimensional representation. 

This principle has two remaining limitations. First, as the number of objects increases, partitioning becomes increasingly difficult, as in autonomous driving tasks with dynamic objects like cars and pedestrians.
 Additionally, the gap between a world model and the real world still needs to be formally quantified to obtain guarantees, which is an open problem for future research.

\section{Experimental Validation}\label{sec:experiments}

The objective of our experiments is to evaluate the impact of the four proposed principles on the interpretability of world model representations. We expect each principle to improve the prediction of future physical states compared to a baseline interpretable world model. The success is measured by 
the mean squared error (MSE) of state predictions over varying prediction horizons. 

Two case studies are used to validate the principles: the \textit{Lunar Lander} and \textit{Cart Pole} environments from OpenAI Gym~\cite[]{brockman2016openaigym}. The state dimensions for the Lunar Lander and Cart Pole are 8 and 4, respectively, reflecting different levels of complexity in achieving interpretability. 
We utilize classical models, namely a Variational Autoencoder (VAE) for encoding observations and a Long Short-Term Memory (LSTM) network for temporal prediction. There are 64 latent dimensions in all experiments. In the baseline interpretable world model, only the first few dimensions are supervised with interpretable physical meanings, whereas the remaining dimensions are not supervised. Additional details are available in the Appendix and \href{https://github.com/Trustworthy-Engineered-Autonomy-Lab/piwm-principles}{online repository}.

  \begin{figure}[h]
	\centering         \includegraphics[width=\textwidth]{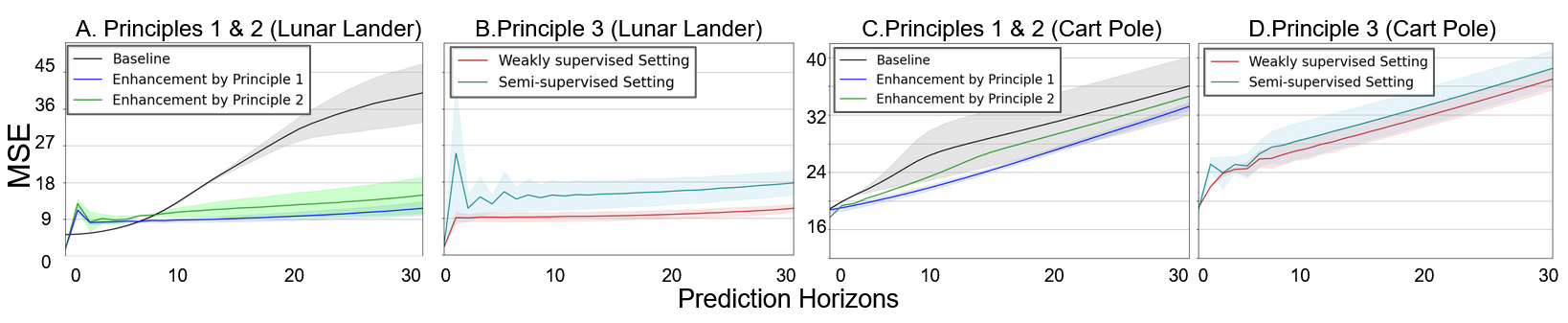}
    \vspace{-9mm}
    \caption{MSE of physical state prediction across different prediction horizons for Principles 1--3.
    }

	\label{fig:p3}
 \end{figure}


\begin{table}[h]
\vspace{-4mm}
    \centering
     \small
    \begin{tabular}{lcccc}
        \hline
        \textbf{World model} & \textbf{Environment} & \textbf{Average MSE $\downarrow$} & \textbf{Average SSIM $\uparrow$} & \textbf{Model Size $\downarrow$} \\
        \hline
        Baseline (monolithic) & Cart Pole & 0.02856 & 0.997122 & 200,259 \\
        Partitioned 3-way & Cart Pole & 0.05176 & 0.995614 & \textbf{144,665} \\
        Baseline (monolithic) & Lunar Lander & 0.18801 & 0.8686 & 360,773 \\
        Partitioned 3-way & Lunar Lander & 0.306 & 0.6289 & \textbf{78,101} \\
        \hline
    \end{tabular}
    \vspace{-2mm}
    \caption{\looseness=-1
    Model size reduction and reconstruction performance for validating Principle 4, $\lambda = 0.2$.}
    \label{tab:comparison}
\end{table}

\noindent
\textbf{Principle 1:} Here we split the encoder into the image part for extracting low-level visual features and the state part that produces values of physical variables. The latent vector size is the same for the baseline and the modified models.
Figures~\ref{fig:p3}A and~\ref{fig:p3}C 
show that Principle 1 significantly reduces the MSE for longer horizons, highlighting the stability that comes from physical interpretability. 

\smallskip
\noindent
\textbf{Principle 2:} We specify a function $g$ that shifts the lunar lander's position, and a corresponding function $g$ that shifts the latent state.
For cart pole, we shift both the rotation and position, with corresponding changes made to the latent state. Figures~\ref{fig:p3}A and \ref{fig:p3}C show that Principle 2 reduces prediction error across all prediction horizons, confirming the value of equivariance. While this principle improves performance on lunar lander, it has less of an effect on the cart pole. We hypothesize that this principle benefits more complex and partially observable systems.

\smallskip
\noindent
\textbf{Principle 3:} Here we train world models in semi- and weakly-supervised settings: (1) only static information (position, angle) is supervised, while dynamic (velocity) is unknown; (2) velocity is estimated from positions/angles, adding supervision through physical knowledge. 
Figures~\ref{fig:p3}B and~\ref{fig:p3}D show that weak physical supervision improves prediction quality at all prediction horizons.

\smallskip
\noindent
\textbf{Principle 4}: We partition the original cartpole and lunar lander images into three parts with SAM, training three smaller decoders for each and combining them as shown in Figure~\ref{fig:parts} in the Appendix. 
The partitioned generator inputs are the exact physical states, 
while the baseline approach's uninterpretable latents. 
Our partitioning reduces the baseline's parameters by \textit{27.7\%} 
while keeping a comparable reconstruction quality remains comparable, as per Table~\ref{tab:comparison}. Though based on simple environments, these standard benchmarks help isolate each principle’s effect.
We plan to extend validation to more complex domains like 3D navigation and visual robotic manipulation.

\section{Future Research Directions}\label{sec:future}

\noindent 
\textbf{A. Extracting Physical Knowledge from Foundation Models.} It is difficult for humans to externalize their implicit knowledge of the physical world~\cite[]{trager_linear_2023,xu_llm-enabled_2024}. Having absorbed humanity-scale data patterns, large language models are promising sources of implicit and plausible physical knowledge. We intend to investigate how to extract candidate dynamics templates, invariances, and equivariances. An important step is validating the candidate information (e.g., via open datasets) before incorporating it into the world model training. 

\smallskip
\looseness=-1
\noindent
\textbf{B. Physically Aligned Multimodality.} Reliable multimodal world models are urgently needed in many autonomous systems~\cite[]{gupta_essential_2024,zheng_neurostrata_2025}. However, the consistency of predicted modalities has been a challenge for learned representations~\cite[]{lu_surveying_2024}. We suggest the use of physically meaningful representations in making image and LiDAR predictions consistent on real-world datasets such as nuPlan~\cite[]{caesar_nuplan_2022} and Waymo Open~\cite[]{sun_scalability_2020}. 

\smallskip
\noindent
\textbf{C. Interpretable Uncertainty in World Models.} Commonplace uncertainty quantification techniques for deep learning models struggle to express the uncertainty in the terms relevant to the application domain~\cite[]{gal2016dropout,kendall2017uncertainties}. Traditional Bayesian approaches and ensemble methods often focus on model uncertainty but fail to capture the structured uncertainty inherent in physical systems~\cite[]{zhang2019quantifying}. In contrast, uncertainty estimation within physically meaningful latent representations allows for more interpretable and actionable uncertainties. We consider it fruitful to develop an uncertainty quantification method based on distributions over physically meaningful latent states and partitioned outputs, which can facilitate robust decision-making and improve reliability in downstream tasks~\cite[]{depeweg2018decomposition}.

\smallskip
\noindent
\textbf{D. Unified Training Pipeline.} We outlined several training objectives and supervision strategies for world models. However, when their combinations are used, the convergence and stability of training remain elusive~\cite[]{sener2018multi}. We recommend developing an automated training pipeline that will combine and tune different losses to ensure reliable training~\cite[]{li2018hyperband}. 

\smallskip
\noindent
\textbf{E. Integrating World Models into Classical Autonomy.} Physically meaningful states enable high-performance components of world models to serve as state estimators, trajectory predictors, and verification models~\cite[]{mao_towards_2024}. This allows combining the previously incompatible first-principles and end-to-end learning models. We intend to improve the performance of classic autonomy tasks with world-model components while preserving their reliability and verifiability.

\acks{
The authors thank Vedansh Maheshwari, Mrinall Umasudhan, Rohith Reddy Nama, Sukanth Sundaran, and Liam Cade McGlothlin for their experiments with neural representations. This research is supported in part by the NSF Grant CCF-2403616. Any opinions, findings, conclusions, or recommendations expressed in this material are those of the authors and do not necessarily reflect the views of the National Science Foundation (NSF) or the United States Government. 
}

\clearpage

\bibliography{bibs/ref,bibs/ivan-autogenerated}

\begin{thebibliography}{112}
\providecommand{\natexlab}[1]{#1}
\providecommand{\url}[1]{\texttt{#1}}
\expandafter\ifx\csname urlstyle\endcsname\relax
  \providecommand{\doi}[1]{doi: #1}\else
  \providecommand{\doi}{doi: \begingroup \urlstyle{rm}\Url}\fi

\bibitem[Alemi et~al.(2018)Alemi, Poole, Fischer, Dillon, Saurous, and Murphy]{alemi2018fixing}
Alexander Alemi, Ben Poole, Ian Fischer, Joshua Dillon, Rif~A Saurous, and Kevin Murphy.
\newblock Fixing a broken elbo.
\newblock In \emph{International conference on machine learning}, pages 159--168. PMLR, 2018.

\bibitem[Althoff(2015)]{Althoff_verify}
Matthias Althoff.
\newblock An introduction to cora 2015.
\newblock In \emph{Proc. of the workshop on applied verification for continuous and hybrid systems}, pages 120--151, 2015.

\bibitem[Baig et~al.(2023)Baig, Ma, Xu, and You]{baig_autoencoder_2023}
Yasa Baig, Helena~R. Ma, Helen Xu, and Lingchong You.
\newblock Autoencoder neural networks enable low dimensional structure analyses of microbial growth dynamics.
\newblock \emph{Nature Communications}, 14\penalty0 (1):\penalty0 7937, December 2023.
\newblock ISSN 2041-1723.
\newblock \doi{10.1038/s41467-023-43455-0}.
\newblock URL \url{https://www.nature.com/articles/s41467-023-43455-0}.
\newblock Publisher: Nature Publishing Group.

\bibitem[Balloch et~al.(2023)Balloch, Lin, Peng, Hussain, Srinivas, Wright, Kim, and Riedl]{balloch_neuro-symbolic_2023}
Jonathan~C. Balloch, Zhiyu Lin, Xiangyu Peng, Mustafa Hussain, Aarun Srinivas, Robert Wright, Julia~M. Kim, and Mark~O. Riedl.
\newblock Neuro-{Symbolic} {World} {Models} for {Adapting} to {Open} {World} {Novelty}.
\newblock In \emph{Proceedings of the 2023 {International} {Conference} on {Autonomous} {Agents} and {Multiagent} {Systems}}, {AAMAS} '23, pages 2848--2850, Richland, SC, May 2023. International Foundation for Autonomous Agents and Multiagent Systems.
\newblock ISBN 978-1-4503-9432-1.

\bibitem[Bar et~al.(2024)Bar, Zhou, Tran, Darrell, and LeCun]{bar_navigation_2024}
Amir Bar, Gaoyue Zhou, Danny Tran, Trevor Darrell, and Yann LeCun.
\newblock Navigation {World} {Models}, December 2024.
\newblock URL \url{http://arxiv.org/abs/2412.03572}.
\newblock arXiv:2412.03572 [cs].

\bibitem[Brockman et~al.(2016)Brockman, Cheung, Pettersson, Schneider, Schulman, Tang, and Zaremba]{brockman2016openaigym}
Greg Brockman, Vicki Cheung, Ludwig Pettersson, Jonas Schneider, John Schulman, Jie Tang, and Wojciech Zaremba.
\newblock Openai gym, 2016.
\newblock URL \url{https://arxiv.org/abs/1606.01540}.

\bibitem[Bruce et~al.(2024)Bruce, Dennis, Edwards, Parker-Holder, Shi, Hughes, Lai, Mavalankar, Steigerwald, Apps, Aytar, Bechtle, Behbahani, Chan, Heess, Gonzalez, Osindero, Ozair, Reed, Zhang, Zolna, Clune, Freitas, Singh, and Rocktäschel]{bruce_genie_2024}
Jake Bruce, Michael Dennis, Ashley Edwards, Jack Parker-Holder, Yuge Shi, Edward Hughes, Matthew Lai, Aditi Mavalankar, Richie Steigerwald, Chris Apps, Yusuf Aytar, Sarah Bechtle, Feryal Behbahani, Stephanie Chan, Nicolas Heess, Lucy Gonzalez, Simon Osindero, Sherjil Ozair, Scott Reed, Jingwei Zhang, Konrad Zolna, Jeff Clune, Nando~de Freitas, Satinder Singh, and Tim Rocktäschel.
\newblock Genie: {Generative} {Interactive} {Environments}, February 2024.
\newblock URL \url{http://arxiv.org/abs/2402.15391}.
\newblock arXiv:2402.15391 [cs].

\bibitem[Brunton et~al.(2016)Brunton, Proctor, and Kutz]{brunton_discovering_2016}
Steven~L. Brunton, Joshua~L. Proctor, and J.~Nathan Kutz.
\newblock Discovering governing equations from data by sparse identification of nonlinear dynamical systems.
\newblock \emph{Proceedings of the National Academy of Sciences of the United States of America}, 113\penalty0 (15):\penalty0 3932--3937, April 2016.
\newblock ISSN 1091-6490.
\newblock \doi{10.1073/pnas.1517384113}.

\bibitem[Caesar et~al.(2022)Caesar, Kabzan, Tan, Fong, Wolff, Lang, Fletcher, Beijbom, and Omari]{caesar_nuplan_2022}
Holger Caesar, Juraj Kabzan, Kok~Seang Tan, Whye~Kit Fong, Eric Wolff, Alex Lang, Luke Fletcher, Oscar Beijbom, and Sammy Omari.
\newblock {NuPlan}: {A} closed-loop {ML}-based planning benchmark for autonomous vehicles, February 2022.
\newblock URL \url{http://arxiv.org/abs/2106.11810}.
\newblock arXiv:2106.11810 [cs].

\bibitem[Cai et~al.(2024)Cai, Fan, and Bak]{chuchu_image_controller}
Feiyang Cai, Chuchu Fan, and Stanley Bak.
\newblock Scalable surrogate verification of image-based neural network control systems using composition and unrolling.
\newblock \emph{arXiv preprint arXiv:2405.18554}, 2024.

\bibitem[Chen et~al.(2022)Chen, Huang, Raghupathi, Chandratreya, Du, and Lipson]{chen_automated_2022}
Boyuan Chen, Kuang Huang, Sunand Raghupathi, Ishaan Chandratreya, Qiang Du, and Hod Lipson.
\newblock Automated discovery of fundamental variables hidden in experimental data.
\newblock \emph{Nature Computational Science}, 2\penalty0 (7):\penalty0 433--442, July 2022.
\newblock ISSN 2662-8457.
\newblock \doi{10.1038/s43588-022-00281-6}.
\newblock URL \url{https://www.nature.com/articles/s43588-022-00281-6}.
\newblock Publisher: Nature Publishing Group.

\bibitem[Chen et~al.(2024)Chen, Wu, Yoon, and Ahn]{chen_transdreamer_2024}
Chang Chen, Yi-Fu Wu, Jaesik Yoon, and Sungjin Ahn.
\newblock {TransDreamer}: {Reinforcement} {Learning} with {Transformer} {World} {Models}, November 2024.
\newblock URL \url{http://arxiv.org/abs/2202.09481}.
\newblock arXiv:2202.09481 [cs].

\bibitem[Chen et~al.(2020)Chen, Kornblith, Norouzi, and Hinton]{chen2020simple}
Ting Chen, Simon Kornblith, Mohammad Norouzi, and Geoffrey Hinton.
\newblock A simple framework for contrastive learning of visual representations.
\newblock In \emph{International conference on machine learning}, pages 1597--1607. PMLR, 2020.

\bibitem[Chen et~al.(2016)Chen, Duan, Houthooft, Schulman, Sutskever, and Abbeel]{chen2016infogan}
Xi~Chen, Yan Duan, Rein Houthooft, John Schulman, Ilya Sutskever, and Pieter Abbeel.
\newblock Infogan: Interpretable representation learning by information maximizing generative adversarial nets.
\newblock \emph{Advances in neural information processing systems}, 29, 2016.

\bibitem[Crawford and Pineau(2020)]{crawford_exploiting_2020}
Eric Crawford and Joelle Pineau.
\newblock Exploiting {Spatial} {Invariance} for {Scalable} {Unsupervised} {Object} {Tracking}.
\newblock \emph{Proceedings of the AAAI Conference on Artificial Intelligence}, 34\penalty0 (04):\penalty0 3684--3692, April 2020.
\newblock ISSN 2374-3468.
\newblock \doi{10.1609/aaai.v34i04.5777}.
\newblock URL \url{https://ojs.aaai.org/index.php/AAAI/article/view/5777}.
\newblock Number: 04.

\bibitem[Deng et~al.(2023)Deng, Park, and Ahn]{deng_facing_2023}
Fei Deng, Junyeong Park, and Sungjin Ahn.
\newblock Facing {Off} {World} {Model} {Backbones}: {RNNs}, {Transformers}, and {S4}.
\newblock In \emph{Proc. of {NeurIPS} 2023}, November 2023.
\newblock \doi{10.48550/arXiv.2307.02064}.
\newblock arXiv:2307.02064 [cs].

\bibitem[Depeweg et~al.(2018)Depeweg, Hernandez-Lobato, Doshi-Velez, and Udluft]{depeweg2018decomposition}
Stefan Depeweg, Jose-Miguel Hernandez-Lobato, Finale Doshi-Velez, and Steffen Udluft.
\newblock Decomposition of uncertainty in bayesian deep learning for efficient and risk-sensitive learning.
\newblock In \emph{International conference on machine learning}, pages 1184--1193. PMLR, 2018.

\bibitem[Ding et~al.(2024)Ding, Zhang, Tian, and Zheng]{ding2024diffusion}
Zihan Ding, Amy Zhang, Yuandong Tian, and Qinqing Zheng.
\newblock Diffusion world model, 2024.

\bibitem[Fremont et~al.(2019)Fremont, Dreossi, Ghosh, Yue, Sangiovanni-Vincentelli, and Seshia]{fremont_scenic_2019}
Daniel~J. Fremont, Tommaso Dreossi, Shromona Ghosh, Xiangyu Yue, Alberto~L. Sangiovanni-Vincentelli, and Sanjit~A. Seshia.
\newblock Scenic: a language for scenario specification and scene generation.
\newblock In \emph{Proceedings of the 40th {ACM} {SIGPLAN} {Conference} on {Programming} {Language} {Design} and {Implementation}}, {PLDI} 2019, pages 63--78, New York, NY, USA, June 2019. Association for Computing Machinery.
\newblock ISBN 978-1-4503-6712-7.
\newblock \doi{10.1145/3314221.3314633}.
\newblock URL \url{https://doi.org/10.1145/3314221.3314633}.

\bibitem[Gal and Ghahramani(2016)]{gal2016dropout}
Yarin Gal and Zoubin Ghahramani.
\newblock Dropout as a bayesian approximation: Representing model uncertainty in deep learning.
\newblock In \emph{international conference on machine learning}, pages 1050--1059. PMLR, 2016.

\bibitem[Garg et~al.(2019)Garg, Chiang, Sugaya, Faust, and Tapia]{garg_comparison_2019}
Arpit Garg, Hao-Tien~Lewis Chiang, Satomi Sugaya, Aleksandra Faust, and Lydia Tapia.
\newblock Comparison of {Deep} {Reinforcement} {Learning} {Policies} to {Formal} {Methods} for {Moving} {Obstacle} {Avoidance}.
\newblock In \emph{2019 {IEEE}/{RSJ} {International} {Conference} on {Intelligent} {Robots} and {Systems} ({IROS})}, pages 3534--3541, November 2019.
\newblock \doi{10.1109/IROS40897.2019.8967945}.
\newblock URL \url{https://ieeexplore.ieee.org/document/8967945}.
\newblock ISSN: 2153-0866.

\bibitem[Ge et~al.(2024)Ge, Huang, Zhou, Li, Wang, Tang, and Zhuang]{ge_worldgpt_2024}
Zhiqi Ge, Hongzhe Huang, Mingze Zhou, Juncheng Li, Guoming Wang, Siliang Tang, and Yueting Zhuang.
\newblock {WorldGPT}: {Empowering} {LLM} as {Multimodal} {World} {Model}, September 2024.
\newblock URL \url{http://arxiv.org/abs/2404.18202}.
\newblock arXiv:2404.18202 [cs].

\bibitem[Geng et~al.(2024)Geng, Dutta, and Ruchkin]{geng2024bridging}
Yuang Geng, Souradeep Dutta, and Ivan Ruchkin.
\newblock Bridging dimensions: Confident reachability for high-dimensional controllers, 2024.

\bibitem[Greydanus et~al.(2019)Greydanus, Dzamba, and Yosinski]{greydanus_hamiltonian_2019}
Samuel Greydanus, Misko Dzamba, and Jason Yosinski.
\newblock Hamiltonian {Neural} {Networks}.
\newblock In \emph{Advances in {Neural} {Information} {Processing} {Systems}}, volume~32. Curran Associates, Inc., 2019.
\newblock URL \url{https://papers.nips.cc/paper_files/paper/2019/hash/26cd8ecadce0d4efd6cc8a8725cbd1f8-Abstract.html}.

\bibitem[Gumbsch et~al.(2023)Gumbsch, Sajid, Martius, and Butz]{gumbsch_learning_2023}
Christian Gumbsch, Noor Sajid, Georg Martius, and Martin~V. Butz.
\newblock Learning {Hierarchical} {World} {Models} with {Adaptive} {Temporal} {Abstractions} from {Discrete} {Latent} {Dynamics}.
\newblock July 2023.
\newblock URL \url{https://openreview.net/forum?id=5qappsbO73r}.

\bibitem[Gupta et~al.(2024)Gupta, Gong, Ma, Pawlowski, Hilmkil, Scetbon, Rigter, Famoti, Llorens, Gao, Bauer, Kragic, Schölkopf, and Zhang]{gupta_essential_2024}
Tarun Gupta, Wenbo Gong, Chao Ma, Nick Pawlowski, Agrin Hilmkil, Meyer Scetbon, Marc Rigter, Ade Famoti, Ashley~Juan Llorens, Jianfeng Gao, Stefan Bauer, Danica Kragic, Bernhard Schölkopf, and Cheng Zhang.
\newblock The {Essential} {Role} of {Causality} in {Foundation} {World} {Models} for {Embodied} {AI}, April 2024.
\newblock URL \url{http://arxiv.org/abs/2402.06665}.
\newblock arXiv:2402.06665.

\bibitem[Ha and Schmidhuber(2018)]{ha_recurrent_2018}
David Ha and Jürgen Schmidhuber.
\newblock Recurrent {World} {Models} {Facilitate} {Policy} {Evolution}.
\newblock In \emph{Advances in {Neural} {Information} {Processing} {Systems}}, volume~31, 2018.

\bibitem[Hafner et~al.(2019)Hafner, Lillicrap, Fischer, Villegas, Ha, Lee, and Davidson]{hafner_learning_2019}
Danijar Hafner, Timothy Lillicrap, Ian Fischer, Ruben Villegas, David Ha, Honglak Lee, and James Davidson.
\newblock Learning {Latent} {Dynamics} for {Planning} from {Pixels}.
\newblock In \emph{Proceedings of the 36th {International} {Conference} on {Machine} {Learning}}, pages 2555--2565. PMLR, May 2019.
\newblock URL \url{https://proceedings.mlr.press/v97/hafner19a.html}.
\newblock ISSN: 2640-3498.

\bibitem[Hafner et~al.(2020)Hafner, Lillicrap, Ba, and Norouzi]{hafner_dream_2020}
Danijar Hafner, Timothy Lillicrap, Jimmy Ba, and Mohammad Norouzi.
\newblock Dream to {Control}: {Learning} {Behaviors} by {Latent} {Imagination}, March 2020.
\newblock URL \url{http://arxiv.org/abs/1912.01603}.
\newblock arXiv:1912.01603 [cs].

\bibitem[Hafner et~al.(2022)Hafner, Lillicrap, Norouzi, and Ba]{hafner2022mastering}
Danijar Hafner, Timothy Lillicrap, Mohammad Norouzi, and Jimmy Ba.
\newblock Mastering atari with discrete world models, 2022.

\bibitem[Hafner et~al.(2024)Hafner, Pasukonis, Ba, and Lillicrap]{hafner_mastering_2024}
Danijar Hafner, Jurgis Pasukonis, Jimmy Ba, and Timothy Lillicrap.
\newblock Mastering {Diverse} {Domains} through {World} {Models}, April 2024.
\newblock URL \url{http://arxiv.org/abs/2301.04104}.
\newblock arXiv:2301.04104 [cs].

\bibitem[Hao et~al.(2023)Hao, Gu, Ma, Hong, Wang, Wang, and Hu]{hao_reasoning_2023}
Shibo Hao, Yi~Gu, Haodi Ma, Joshua~Jiahua Hong, Zhen Wang, Daisy~Zhe Wang, and Zhiting Hu.
\newblock Reasoning with {Language} {Model} is {Planning} with {World} {Model}, October 2023.
\newblock URL \url{http://arxiv.org/abs/2305.14992}.
\newblock arXiv:2305.14992 [cs].

\bibitem[Higgins et~al.(2016)Higgins, Matthey, Pal, Burgess, Glorot, Botvinick, Mohamed, and Lerchner]{higgins_beta-vae_2016}
Irina Higgins, Loic Matthey, Arka Pal, Christopher Burgess, Xavier Glorot, Matthew Botvinick, Shakir Mohamed, and Alexander Lerchner.
\newblock beta-{VAE}: {Learning} {Basic} {Visual} {Concepts} with a {Constrained} {Variational} {Framework}.
\newblock November 2016.
\newblock URL \url{https://openreview.net/forum?id=Sy2fzU9gl}.

\bibitem[Hinton(1986)]{hinton_learning_1986}
Geoffrey~E. Hinton.
\newblock Learning {Distributed} {Representations} of {Concepts}.
\newblock \emph{Proceedings of the Annual Meeting of the Cognitive Science Society}, 8\penalty0 (0), 1986.
\newblock URL \url{https://escholarship.org/uc/item/79w838g1}.

\bibitem[Hu et~al.(2023)Hu, Russell, Yeo, Murez, Fedoseev, Kendall, Shotton, and Corrado]{hu_gaia-1_2023}
Anthony Hu, Lloyd Russell, Hudson Yeo, Zak Murez, George Fedoseev, Alex Kendall, Jamie Shotton, and Gianluca Corrado.
\newblock {GAIA}-1: {A} {Generative} {World} {Model} for {Autonomous} {Driving}, September 2023.
\newblock URL \url{http://arxiv.org/abs/2309.17080}.
\newblock arXiv:2309.17080 [cs].

\bibitem[Itkina and Kochenderfer(2022)]{itkina_interpretable_2022}
Masha Itkina and Mykel Kochenderfer.
\newblock Interpretable {Self}-{Aware} {Neural} {Networks} for {Robust} {Trajectory} {Prediction}.
\newblock August 2022.
\newblock URL \url{https://openreview.net/forum?id=fnaMlJbRc4t}.

\bibitem[Jiang et~al.(2020)Jiang, Janghorbani, Melo, and Ahn]{jiang_scalor_2020}
Jindong Jiang, Sepehr Janghorbani, Gerard~de Melo, and Sungjin Ahn.
\newblock {SCALOR}: {Generative} {World} {Models} with {Scalable} {Object} {Representations}, March 2020.
\newblock URL \url{http://arxiv.org/abs/1910.02384}.
\newblock arXiv:1910.02384 [cs].

\bibitem[Kalman(1960)]{kalman1960new}
Rudolph~Emil Kalman.
\newblock A new approach to linear filtering and prediction problems.
\newblock 1960.

\bibitem[Katz et~al.(2022)Katz, Corso, Strong, and Kochenderfer]{katz_verification_2022}
Sydney~M. Katz, Anthony~L. Corso, Christopher~A. Strong, and Mykel~J. Kochenderfer.
\newblock Verification of {Image}-{Based} {Neural} {Network} {Controllers} {Using} {Generative} {Models}.
\newblock \emph{Journal of Aerospace Information Systems}, 19\penalty0 (9):\penalty0 574--584, 2022.
\newblock ISSN 1940-3151.
\newblock \doi{10.2514/1.I011071}.
\newblock URL \url{https://doi.org/10.2514/1.I011071}.
\newblock Publisher: American Institute of Aeronautics and Astronautics \_eprint: https://doi.org/10.2514/1.I011071.

\bibitem[Kendall and Gal(2017)]{kendall2017uncertainties}
Alex Kendall and Yarin Gal.
\newblock What uncertainties do we need in bayesian deep learning for computer vision?
\newblock \emph{Advances in neural information processing systems}, 30, 2017.

\bibitem[Kim et~al.(2020)Kim, Sano, Freitas, Haber, and Yamins]{kim_active_2020}
Kuno Kim, Megumi Sano, Julian~De Freitas, Nick Haber, and Daniel Yamins.
\newblock Active {World} {Model} {Learning} with {Progress} {Curiosity}, July 2020.
\newblock URL \url{http://arxiv.org/abs/2007.07853}.
\newblock arXiv:2007.07853 [cs].

\bibitem[Kipf et~al.(2018)Kipf, Fetaya, Wang, Welling, and Zemel]{kipf_neural_2018}
Thomas Kipf, Ethan Fetaya, Kuan-Chieh Wang, Max Welling, and Richard Zemel.
\newblock Neural {Relational} {Inference} for {Interacting} {Systems}, June 2018.
\newblock URL \url{http://arxiv.org/abs/1802.04687}.

\bibitem[Kirillov et~al.(2023)Kirillov, Mintun, Ravi, Mao, Rolland, Gustafson, Xiao, Whitehead, Berg, Lo, et~al.]{SAE}
Alexander Kirillov, Eric Mintun, Nikhila Ravi, Hanzi Mao, Chloe Rolland, Laura Gustafson, Tete Xiao, Spencer Whitehead, Alexander~C Berg, Wan-Yen Lo, et~al.
\newblock Segment anything.
\newblock In \emph{Proceedings of the IEEE/CVF International Conference on Computer Vision}, pages 4015--4026, 2023.

\bibitem[Koh et~al.(2021)Koh, Lee, Yang, Baldridge, and Anderson]{koh_pathdreamer_2021}
Jing~Yu Koh, Honglak Lee, Yinfei Yang, Jason Baldridge, and Peter Anderson.
\newblock Pathdreamer: {A} {World} {Model} for {Indoor} {Navigation}, August 2021.
\newblock URL \url{http://arxiv.org/abs/2105.08756}.
\newblock arXiv:2105.08756 [cs].

\bibitem[Kolev et~al.(2024)Kolev, Rafailov, Hatch, Wu, and Finn]{kolev_efficient_2024}
Victor Kolev, Rafael Rafailov, Kyle Hatch, Jiajun Wu, and Chelsea Finn.
\newblock Efficient {Imitation} {Learning} with {Conservative} {World} {Models}, August 2024.
\newblock URL \url{http://arxiv.org/abs/2405.13193}.
\newblock arXiv:2405.13193 [cs].

\bibitem[Kosiorek et~al.(2018)Kosiorek, Kim, Posner, and Teh]{kosiorek_sequential_2018}
Adam~R. Kosiorek, Hyunjik Kim, Ingmar Posner, and Yee~Whye Teh.
\newblock Sequential {Attend}, {Infer}, {Repeat}: {Generative} {Modelling} of {Moving} {Objects}, November 2018.
\newblock URL \url{http://arxiv.org/abs/1806.01794}.
\newblock arXiv:1806.01794 [cs].

\bibitem[Kossen et~al.(2020)Kossen, Stelzner, Hussing, Voelcker, and Kersting]{kossen_structured_2020}
Jannik Kossen, Karl Stelzner, Marcel Hussing, Claas Voelcker, and Kristian Kersting.
\newblock Structured {Object}-{Aware} {Physics} {Prediction} for {Video} {Modeling} and {Planning}, February 2020.
\newblock URL \url{http://arxiv.org/abs/1910.02425}.
\newblock arXiv:1910.02425 [cs].

\bibitem[Lee et~al.(2013)]{lee2013pseudo}
Dong-Hyun Lee et~al.
\newblock Pseudo-label: The simple and efficient semi-supervised learning method for deep neural networks.
\newblock In \emph{Workshop on challenges in representation learning, ICML}, volume~3, page 896. Atlanta, 2013.

\bibitem[Li et~al.(2018)Li, Jamieson, DeSalvo, Rostamizadeh, and Talwalkar]{li2018hyperband}
Lisha Li, Kevin Jamieson, Giulia DeSalvo, Afshin Rostamizadeh, and Ameet Talwalkar.
\newblock Hyperband: A novel bandit-based approach to hyperparameter optimization.
\newblock \emph{Journal of Machine Learning Research}, 18\penalty0 (185):\penalty0 1--52, 2018.

\bibitem[Li et~al.(2024)Li, Jia, Wang, and Yan]{li_think2drive_2024}
Qifeng Li, Xiaosong Jia, Shaobo Wang, and Junchi Yan.
\newblock {Think2Drive}: {Efficient} {Reinforcement} {Learning} by {Thinking} in {Latent} {World} {Model} for {Quasi}-{Realistic} {Autonomous} {Driving} (in {CARLA}-v2), July 2024.
\newblock URL \url{http://arxiv.org/abs/2402.16720}.
\newblock arXiv:2402.16720 [cs].

\bibitem[Liang et~al.(2024)Liang, Kumar, Tang, Weller, Tenenbaum, Silver, Henriques, and Ellis]{liang_visualpredicator_2024}
Yichao Liang, Nishanth Kumar, Hao Tang, Adrian Weller, Joshua~B. Tenenbaum, Tom Silver, João~F. Henriques, and Kevin Ellis.
\newblock {VisualPredicator}: {Learning} {Abstract} {World} {Models} with {Neuro}-{Symbolic} {Predicates} for {Robot} {Planning}, October 2024.
\newblock URL \url{http://arxiv.org/abs/2410.23156}.
\newblock arXiv:2410.23156 [cs].

\bibitem[Lin et~al.(2020)Lin, Wu, Peri, Fu, Jiang, and Ahn]{lin_improving_2020}
Zhixuan Lin, Yi-Fu Wu, Skand Peri, Bofeng Fu, Jindong Jiang, and Sungjin Ahn.
\newblock Improving {Generative} {Imagination} in {Object}-{Centric} {World} {Models}, October 2020.
\newblock URL \url{http://arxiv.org/abs/2010.02054}.
\newblock arXiv:2010.02054 [cs].

\bibitem[Linial et~al.(2021)Linial, Ravid, Eytan, and Shalit]{linial_generative_2021}
Ori Linial, Neta Ravid, Danny Eytan, and Uri Shalit.
\newblock Generative {ODE} modeling with known unknowns.
\newblock In \emph{Proceedings of the {Conference} on {Health}, {Inference}, and {Learning}}, {CHIL} '21, pages 79--94, New York, NY, USA, April 2021. Association for Computing Machinery.
\newblock ISBN 978-1-4503-8359-2.
\newblock \doi{10.1145/3450439.3451866}.
\newblock URL \url{https://dl.acm.org/doi/10.1145/3450439.3451866}.

\bibitem[Liu et~al.(2023)Liu, Xiao, and Belta]{liu_learning_2023}
Wenliang Liu, Wei Xiao, and Calin Belta.
\newblock Learning {Robust} and {Correct} {Controllers} from {Signal} {Temporal} {Logic} {Specifications} {Using} {BarrierNet}.
\newblock 2023.
\newblock \doi{10.48550/ARXIV.2304.06160}.
\newblock URL \url{https://arxiv.org/abs/2304.06160}.
\newblock Publisher: arXiv Version Number: 1.

\bibitem[Lu et~al.(2024{\natexlab{a}})Lu, Zhan, Tomizuka, and Hu]{lu_towards_2024}
Juanwu Lu, Wei Zhan, Masayoshi Tomizuka, and Yeping Hu.
\newblock Towards {Generalizable} and {Interpretable} {Motion} {Prediction}: {A} {Deep} {Variational} {Bayes} {Approach}.
\newblock In \emph{Proceedings of {The} 27th {International} {Conference} on {Artificial} {Intelligence} and {Statistics}}, pages 4717--4725. PMLR, April 2024{\natexlab{a}}.
\newblock URL \url{https://proceedings.mlr.press/v238/lu24a.html}.
\newblock ISSN: 2640-3498.

\bibitem[Lu et~al.(2024{\natexlab{b}})Lu, Afridi, Kang, Ruchkin, and Zheng]{lu_surveying_2024}
Zhen Lu, Imran Afridi, Hong~Jin Kang, Ivan Ruchkin, and Xi~Zheng.
\newblock Surveying neuro-symbolic approaches for reliable artificial intelligence of things.
\newblock \emph{Journal of Reliable Intelligent Environments}, July 2024{\natexlab{b}}.
\newblock ISSN 2199-4676.
\newblock \doi{10.1007/s40860-024-00231-1}.
\newblock URL \url{https://doi.org/10.1007/s40860-024-00231-1}.

\bibitem[Ma et~al.(2024)Ma, Wu, Feng, Xiao, Li, Hao, Wang, and Long]{ma_harmonydream_2024}
Haoyu Ma, Jialong Wu, Ningya Feng, Chenjun Xiao, Dong Li, Jianye Hao, Jianmin Wang, and Mingsheng Long.
\newblock {HarmonyDream}: {Task} {Harmonization} {Inside} {World} {Models}.
\newblock June 2024.
\newblock URL \url{https://openreview.net/forum?id=x0yIaw2fgk}.

\bibitem[Mao et~al.(2025)Mao, Gu, Sha, Shao, Wang, and Abdelzaher]{mao_phy-taylor_2025}
Yanbing Mao, Yuliang Gu, Lui Sha, Huajie Shao, Qixin Wang, and Tarek Abdelzaher.
\newblock Phy-{Taylor}: {Partially} {Physics}-{Knowledge}-{Enhanced} {Deep} {Neural} {Networks} via {NN} {Editing}.
\newblock \emph{IEEE Transactions on Neural Networks and Learning Systems}, 36\penalty0 (1):\penalty0 447--461, January 2025.
\newblock ISSN 2162-2388.
\newblock \doi{10.1109/TNNLS.2023.3325432}.
\newblock URL \url{https://ieeexplore.ieee.org/document/10297119}.
\newblock Conference Name: IEEE Transactions on Neural Networks and Learning Systems.

\bibitem[Mao and Ruchkin(2024)]{mao_towards_2024}
Zhenjiang Mao and Ivan Ruchkin.
\newblock Towards {Physically} {Interpretable} {World} {Models}: {Meaningful} {Weakly} {Supervised} {Representations} for {Visual} {Trajectory} {Prediction}, December 2024.
\newblock URL \url{http://arxiv.org/abs/2412.12870}.
\newblock arXiv:2412.12870 [cs].

\bibitem[Mao et~al.(2024{\natexlab{a}})Mao, Dai, Geng, and Ruchkin]{mao_zero-shot_2024}
Zhenjiang Mao, Siqi Dai, Yuang Geng, and Ivan Ruchkin.
\newblock Zero-shot {Safety} {Prediction} for {Autonomous} {Robots} with {Foundation} {World} {Models}.
\newblock In \emph{Back to the {Future}: {Robot} {Learning} {Going} {Probabilistic} {Workshop} (co-located with {ICRA} 2024)}, March 2024{\natexlab{a}}.
\newblock \doi{10.48550/arXiv.2404.00462}.
\newblock URL \url{http://arxiv.org/abs/2404.00462}.
\newblock arXiv:2404.00462 [cs] version: 1.

\bibitem[Mao et~al.(2024{\natexlab{b}})Mao, Sobolewski, and Ruchkin]{mao_how_2024}
Zhenjiang Mao, Carson Sobolewski, and Ivan Ruchkin.
\newblock How {Safe} {Am} {I} {Given} {What} {I} {See}? {Calibrated} {Prediction} of {Safety} {Chances} for {Image}-{Controlled} {Autonomy}.
\newblock In \emph{Proc. of the {Annual} {Conference} on {Learning} for {Dynamics} and {Control} ({L4DC})}, 2024{\natexlab{b}}.
\newblock \doi{10.48550/arXiv.2308.12252}.
\newblock URL \url{http://arxiv.org/abs/2308.12252}.
\newblock arXiv:2308.12252 [cs].

\bibitem[Mendonca et~al.(2023)Mendonca, Bahl, and Pathak]{mendonca_structured_2023}
Russell Mendonca, Shikhar Bahl, and Deepak Pathak.
\newblock Structured {World} {Models} from {Human} {Videos}, August 2023.
\newblock URL \url{http://arxiv.org/abs/2308.10901}.
\newblock arXiv:2308.10901 [cs].

\bibitem[Miao et~al.(2025)Miao, Fainekos, Hoxha, Okamoto, Prokhorov, and Mitra]{miao_dashcam_2025}
Yan Miao, Georgios Fainekos, Bardh Hoxha, Hideki Okamoto, Danil Prokhorov, and Sayan Mitra.
\newblock From {Dashcam} {Videos} to {Driving} {Simulations}: {Stress} {Testing} {Automated} {Vehicles} against {Rare} {Events}, January 2025.
\newblock URL \url{http://arxiv.org/abs/2411.16027}.
\newblock arXiv:2411.16027 [cs].

\bibitem[Micheli et~al.(2023)Micheli, Alonso, and Fleuret]{micheli_transformers_2023}
Vincent Micheli, Eloi Alonso, and François Fleuret.
\newblock Transformers are {Sample}-{Efficient} {World} {Models}.
\newblock 2023.
\newblock URL \url{https://openreview.net/forum?id=vhFu1Acb0xb}.

\bibitem[Min et~al.(2023)Min, Zhao, Xiao, Nie, and Dai]{min_uniworld_2023}
Chen Min, Dawei Zhao, Liang Xiao, Yiming Nie, and Bin Dai.
\newblock {UniWorld}: {Autonomous} {Driving} {Pre}-training via {World} {Models}, August 2023.
\newblock URL \url{http://arxiv.org/abs/2308.07234}.
\newblock arXiv:2308.07234 [cs].

\bibitem[Min et~al.(2024)Min, Zhao, Xiao, Zhao, Xu, Zhu, Jin, Li, Guo, Xing, Jing, Nie, and Dai]{min_driveworld_2024}
Chen Min, Dawei Zhao, Liang Xiao, Jian Zhao, Xinli Xu, Zheng Zhu, Lei Jin, Jianshu Li, Yulan Guo, Junliang Xing, Liping Jing, Yiming Nie, and Bin Dai.
\newblock {DriveWorld}: {4D} {Pre}-trained {Scene} {Understanding} via {World} {Models} for {Autonomous} {Driving}, May 2024.
\newblock URL \url{http://arxiv.org/abs/2405.04390}.
\newblock arXiv:2405.04390 [cs].

\bibitem[Nakano et~al.(2022)Nakano, Suzuki, and Matsuo]{nakano_interaction-based_2022}
Akihiro Nakano, Masahiro Suzuki, and Yutaka Matsuo.
\newblock Interaction-{Based} {Disentanglement} of {Entities} for {Object}-{Centric} {World} {Models}.
\newblock September 2022.
\newblock URL \url{https://openreview.net/forum?id=JQc2VowqCzz}.

\bibitem[Nguyen et~al.(2024)Nguyen, Yang, Buckley, and Ororbia]{nguyen_r-aif_2024}
Viet~Dung Nguyen, Zhizhuo Yang, Christopher~L. Buckley, and Alexander Ororbia.
\newblock R-{AIF}: {Solving} {Sparse}-{Reward} {Robotic} {Tasks} from {Pixels} with {Active} {Inference} and {World} {Models}, September 2024.
\newblock URL \url{http://arxiv.org/abs/2409.14216}.
\newblock arXiv:2409.14216 [cs].

\bibitem[Okada and Taniguchi(2022)]{okada_dreamingv2_2022}
Masashi Okada and Tadahiro Taniguchi.
\newblock {DreamingV2}: {Reinforcement} {Learning} with {Discrete} {World} {Models} without {Reconstruction}, March 2022.
\newblock URL \url{http://arxiv.org/abs/2203.00494}.
\newblock arXiv:2203.00494 [cs].

\bibitem[Park et~al.(2022)Park, Biza, Zhao, Meent, and Walters]{park_learning_2022}
Jung~Yeon Park, Ondrej Biza, Linfeng Zhao, Jan Willem van~de Meent, and Robin Walters.
\newblock Learning {Symmetric} {Embeddings} for {Equivariant} {World} {Models}, June 2022.
\newblock URL \url{http://arxiv.org/abs/2204.11371}.
\newblock arXiv:2204.11371 [cs].

\bibitem[P{\u{a}}s{\u{a}}reanu et~al.(2023)P{\u{a}}s{\u{a}}reanu, Mangal, Gopinath, Getir~Yaman, Imrie, Calinescu, and Yu]{closed_loop_verify}
Corina~S P{\u{a}}s{\u{a}}reanu, Ravi Mangal, Divya Gopinath, Sinem Getir~Yaman, Calum Imrie, Radu Calinescu, and Huafeng Yu.
\newblock Closed-loop analysis of vision-based autonomous systems: A case study.
\newblock In \emph{International conference on computer aided verification}, pages 289--303. Springer, 2023.

\bibitem[Pol et~al.(2020)Pol, Kipf, Oliehoek, and Welling]{pol_plannable_2020}
Elise van~der Pol, Thomas Kipf, Frans~A. Oliehoek, and Max Welling.
\newblock Plannable {Approximations} to {MDP} {Homomorphisms}: {Equivariance} under {Actions}, February 2020.
\newblock URL \url{http://arxiv.org/abs/2002.11963}.
\newblock arXiv:2002.11963 [cs].

\bibitem[Popov et~al.(2024)Popov, Degirmenci, Wehr, Hegde, Oldja, Kamenev, Douillard, Nistér, Muller, Bhargava, Birchfield, and Smolyanskiy]{popov_mitigating_2024}
Alexander Popov, Alperen Degirmenci, David Wehr, Shashank Hegde, Ryan Oldja, Alexey Kamenev, Bertrand Douillard, David Nistér, Urs Muller, Ruchi Bhargava, Stan Birchfield, and Nikolai Smolyanskiy.
\newblock Mitigating {Covariate} {Shift} in {Imitation} {Learning} for {Autonomous} {Vehicles} {Using} {Latent} {Space} {Generative} {World} {Models}, September 2024.
\newblock URL \url{http://arxiv.org/abs/2409.16663}.
\newblock arXiv:2409.16663 [cs].

\bibitem[Poudel et~al.(2022)Poudel, Pandya, and Cipolla]{poudel2022contrastive}
Rudra~PK Poudel, Harit Pandya, and Roberto Cipolla.
\newblock Contrastive unsupervised learning of world model with invariant causal features.
\newblock \emph{arXiv preprint arXiv:2209.14932}, 2022.

\bibitem[Raissi et~al.(2019)Raissi, Perdikaris, and Karniadakis]{raissi_physics-informed_2019}
M.~Raissi, P.~Perdikaris, and G.~E. Karniadakis.
\newblock Physics-informed neural networks: {A} deep learning framework for solving forward and inverse problems involving nonlinear partial differential equations.
\newblock \emph{Journal of Computational Physics}, 378:\penalty0 686--707, February 2019.
\newblock ISSN 0021-9991.
\newblock \doi{10.1016/j.jcp.2018.10.045}.
\newblock URL \url{https://www.sciencedirect.com/science/article/pii/S0021999118307125}.

\bibitem[Rezende et~al.(2014)Rezende, Mohamed, and Wierstra]{rezende2014stochastic}
Danilo~Jimenez Rezende, Shakir Mohamed, and Daan Wierstra.
\newblock Stochastic backpropagation and approximate inference in deep generative models.
\newblock In \emph{International conference on machine learning}, pages 1278--1286. PMLR, 2014.

\bibitem[Rigter et~al.(2024)Rigter, Gupta, Hilmkil, and Ma]{rigter_avid_2024}
Marc Rigter, Tarun Gupta, Agrin Hilmkil, and Chao Ma.
\newblock {AVID}: {Adapting} {Video} {Diffusion} {Models} to {World} {Models}, November 2024.
\newblock URL \url{http://arxiv.org/abs/2410.12822}.
\newblock arXiv:2410.12822 [cs].

\bibitem[Robine et~al.(2023)Robine, Höftmann, Uelwer, and Harmeling]{robine_transformer-based_2023}
Jan Robine, Marc Höftmann, Tobias Uelwer, and Stefan Harmeling.
\newblock Transformer-based {World} {Models} {Are} {Happy} {With} 100k {Interactions}.
\newblock 2023.
\newblock URL \url{https://openreview.net/forum?id=TdBaDGCpjly}.

\bibitem[Saemundsson et~al.(2020)Saemundsson, Terenin, Hofmann, and Deisenroth]{saemundsson_variational_2020}
Steindor Saemundsson, Alexander Terenin, Katja Hofmann, and Marc Deisenroth.
\newblock Variational {Integrator} {Networks} for {Physically} {Structured} {Embeddings}.
\newblock In \emph{Proceedings of the {Twenty} {Third} {International} {Conference} on {Artificial} {Intelligence} and {Statistics}}, pages 3078--3087. PMLR, June 2020.
\newblock URL \url{https://proceedings.mlr.press/v108/saemundsson20a.html}.
\newblock ISSN: 2640-3498.

\bibitem[Saidi et~al.(2022)Saidi, Ziegenbein, Deshmukh, and Ernst]{saidi_autonomous_2022}
Selma Saidi, Dirk Ziegenbein, Jyotirmoy~V. Deshmukh, and Rolf Ernst.
\newblock Autonomous {Systems} {Design}: {Charting} a {New} {Discipline}.
\newblock \emph{IEEE Design \& Test}, 39\penalty0 (1):\penalty0 8--23, February 2022.
\newblock ISSN 2168-2364.
\newblock \doi{10.1109/MDAT.2021.3128434}.
\newblock Conference Name: IEEE Design \& Test.

\bibitem[Sekar et~al.(2020)Sekar, Rybkin, Daniilidis, Abbeel, Hafner, and Pathak]{sekar2020planning}
Ramanan Sekar, Oleh Rybkin, Kostas Daniilidis, Pieter Abbeel, Danijar Hafner, and Deepak Pathak.
\newblock Planning to explore via self-supervised world models.
\newblock In \emph{International conference on machine learning}, pages 8583--8592. PMLR, 2020.

\bibitem[Sener and Koltun(2018)]{sener2018multi}
Ozan Sener and Vladlen Koltun.
\newblock Multi-task learning as multi-objective optimization.
\newblock \emph{Advances in neural information processing systems}, 31, 2018.

\bibitem[Seo et~al.(2023)Seo, Hafner, Liu, Liu, James, Lee, and Abbeel]{seo_masked_2023}
Younggyo Seo, Danijar Hafner, Hao Liu, Fangchen Liu, Stephen James, Kimin Lee, and Pieter Abbeel.
\newblock Masked {World} {Models} for {Visual} {Control}, May 2023.
\newblock URL \url{http://arxiv.org/abs/2206.14244}.
\newblock arXiv:2206.14244 [cs].

\bibitem[Shaj et~al.()Shaj, Zadeh, Demir, Douat, and Neumann]{shaj_multi_nodate}
Vaisakh Shaj, Saleh~Gholam Zadeh, Ozan Demir, Luiz~Ricardo Douat, and Gerhard Neumann.
\newblock Multi {Time} {Scale} {World} {Models}.

\bibitem[Sohn et~al.(2020)Sohn, Berthelot, Carlini, Zhang, Zhang, Raffel, Cubuk, Kurakin, and Li]{sohn2020fixmatch}
Kihyuk Sohn, David Berthelot, Nicholas Carlini, Zizhao Zhang, Han Zhang, Colin~A Raffel, Ekin~Dogus Cubuk, Alexey Kurakin, and Chun-Liang Li.
\newblock Fixmatch: Simplifying semi-supervised learning with consistency and confidence.
\newblock \emph{Advances in neural information processing systems}, 33:\penalty0 596--608, 2020.

\bibitem[Sorokin and Gurevych(2017)]{sorokin2017end}
Daniil Sorokin and Iryna Gurevych.
\newblock End-to-end representation learning for question answering with weak supervision.
\newblock In \emph{Semantic Web Challenges: 4th SemWebEval Challenge at ESWC 2017, Portoroz, Slovenia, May 28-June 1, 2017, Revised Selected Papers}, pages 70--83. Springer, 2017.

\bibitem[Sridhar et~al.(2023)Sridhar, Dutta, Weimer, and Lee]{sridhar_guaranteed_2023}
Kaustubh Sridhar, Souradeep Dutta, James Weimer, and Insup Lee.
\newblock Guaranteed {Conformance} of {Neurosymbolic} {Models} to {Natural} {Constraints}.
\newblock In \emph{{L4DC} 2023}, April 2023.
\newblock \doi{10.48550/arXiv.2212.01346}.
\newblock URL \url{http://arxiv.org/abs/2212.01346}.
\newblock arXiv:2212.01346 [cs].

\bibitem[Sun et~al.(2020)Sun, Kretzschmar, Dotiwalla, Chouard, Patnaik, Tsui, Guo, Zhou, Chai, Caine, Vasudevan, Han, Ngiam, Zhao, Timofeev, Ettinger, Krivokon, Gao, Joshi, Zhang, Shlens, Chen, and Anguelov]{sun_scalability_2020}
Pei Sun, Henrik Kretzschmar, Xerxes Dotiwalla, Aurélien Chouard, Vijaysai Patnaik, Paul Tsui, James Guo, Yin Zhou, Yuning Chai, Benjamin Caine, Vijay Vasudevan, Wei Han, Jiquan Ngiam, Hang Zhao, Aleksei Timofeev, Scott Ettinger, Maxim Krivokon, Amy Gao, Aditya Joshi, Yu~Zhang, Jonathon Shlens, Zhifeng Chen, and Dragomir Anguelov.
\newblock Scalability in {Perception} for {Autonomous} {Driving}: {Waymo} {Open} {Dataset}.
\newblock In \emph{2020 {IEEE}/{CVF} {Conference} on {Computer} {Vision} and {Pattern} {Recognition} ({CVPR})}, pages 2443--2451, June 2020.
\newblock \doi{10.1109/CVPR42600.2020.00252}.
\newblock URL \url{https://ieeexplore.ieee.org/document/9156973}.
\newblock ISSN: 2575-7075.

\bibitem[Tanwani et~al.(2020)Tanwani, Sermanet, Yan, Anand, Phielipp, and Goldberg]{tanwani2020motion2vec}
Ajay~Kumar Tanwani, Pierre Sermanet, Andy Yan, Raghav Anand, Mariano Phielipp, and Ken Goldberg.
\newblock Motion2vec: Semi-supervised representation learning from surgical videos.
\newblock In \emph{2020 IEEE International Conference on Robotics and Automation (ICRA)}, pages 2174--2181. IEEE, 2020.

\bibitem[Tarvainen and Valpola(2017)]{tarvainen2017mean}
Antti Tarvainen and Harri Valpola.
\newblock Mean teachers are better role models: Weight-averaged consistency targets improve semi-supervised deep learning results.
\newblock \emph{Advances in neural information processing systems}, 30, 2017.

\bibitem[Teeti et~al.(2022)Teeti, Khan, Shahbaz, Bradley, Cuzzolin, and De~Raedt]{vision_trajectoy}
Izzeddin Teeti, Salman Khan, Ajmal Shahbaz, Andrew Bradley, Fabio Cuzzolin, and Lud De~Raedt.
\newblock Vision-based intention and trajectory prediction in autonomous vehicles: A survey.
\newblock In \emph{IJCAI}, pages 5630--5637, 2022.

\bibitem[Topcu et~al.(2020)Topcu, Bliss, Cooke, Cummings, Llorens, Shrobe, and Zuck]{topcu_assured_2020}
Ufuk Topcu, Nadya Bliss, Nancy Cooke, Missy Cummings, Ashley Llorens, Howard Shrobe, and Lenore Zuck.
\newblock Assured {Autonomy}: {Path} {Toward} {Living} {With} {Autonomous} {Systems} {We} {Can} {Trust}, October 2020.
\newblock URL \url{http://arxiv.org/abs/2010.14443}.
\newblock arXiv:2010.14443 [cs].

\bibitem[Trager et~al.(2023)Trager, Perera, Zancato, Achille, Bhatia, and Soatto]{trager_linear_2023}
Matthew Trager, Pramuditha Perera, Luca Zancato, Alessandro Achille, Parminder Bhatia, and Stefano Soatto.
\newblock Linear {Spaces} of {Meanings}: {Compositional} {Structures} in {Vision}-{Language} {Models}.
\newblock \emph{2023 IEEE/CVF International Conference on Computer Vision (ICCV)}, pages 15349--15358, October 2023.
\newblock \doi{10.1109/ICCV51070.2023.01412}.
\newblock URL \url{https://ieeexplore.ieee.org/document/10377972/}.
\newblock Conference Name: 2023 IEEE/CVF International Conference on Computer Vision (ICCV) ISBN: 9798350307184 Place: Paris, France Publisher: IEEE.

\bibitem[Tumu et~al.(2023)Tumu, Lindemann, Nghiem, and Mangharam]{tumu_physics_2023}
Renukanandan Tumu, Lars Lindemann, Truong Nghiem, and Rahul Mangharam.
\newblock Physics {Constrained} {Motion} {Prediction} with {Uncertainty} {Quantification}.
\newblock In \emph{Intelligent {Vehicles} 2023}. arXiv, May 2023.
\newblock URL \url{http://arxiv.org/abs/2302.01060}.
\newblock arXiv:2302.01060 [cs].

\bibitem[Wang et~al.(2023{\natexlab{a}})Wang, Du, Torralba, Isola, Zhang, and Tian]{wang_denoised_2023}
Tongzhou Wang, Simon~S. Du, Antonio Torralba, Phillip Isola, Amy Zhang, and Yuandong Tian.
\newblock Denoised {MDPs}: {Learning} {World} {Models} {Better} {Than} the {World} {Itself}, April 2023{\natexlab{a}}.
\newblock URL \url{http://arxiv.org/abs/2206.15477}.
\newblock arXiv:2206.15477 [cs].

\bibitem[Wang et~al.(2023{\natexlab{b}})Wang, Zhu, Huang, Chen, Zhu, and Lu]{wang_drivedreamer_2023}
Xiaofeng Wang, Zheng Zhu, Guan Huang, Xinze Chen, Jiagang Zhu, and Jiwen Lu.
\newblock {DriveDreamer}: {Towards} {Real}-world-driven {World} {Models} for {Autonomous} {Driving}, November 2023{\natexlab{b}}.
\newblock URL \url{http://arxiv.org/abs/2309.09777}.
\newblock arXiv:2309.09777 [cs].

\bibitem[Wei et~al.(2024)Wei, Yuan, Li, Hu, Gan, and Ding]{wei_occllama_2024}
Julong Wei, Shanshuai Yuan, Pengfei Li, Qingda Hu, Zhongxue Gan, and Wenchao Ding.
\newblock {OccLLaMA}: {An} {Occupancy}-{Language}-{Action} {Generative} {World} {Model} for {Autonomous} {Driving}, September 2024.
\newblock URL \url{http://arxiv.org/abs/2409.03272}.
\newblock arXiv:2409.03272 [cs].

\bibitem[Wong et~al.(2023)Wong, Grand, Lew, Goodman, Mansinghka, Andreas, and Tenenbaum]{wong_word_2023}
Lionel Wong, Gabriel Grand, Alexander~K. Lew, Noah~D. Goodman, Vikash~K. Mansinghka, Jacob Andreas, and Joshua~B. Tenenbaum.
\newblock From {Word} {Models} to {World} {Models}: {Translating} from {Natural} {Language} to the {Probabilistic} {Language} of {Thought}, June 2023.
\newblock URL \url{http://arxiv.org/abs/2306.12672}.
\newblock arXiv:2306.12672 [cs].

\bibitem[Wu et~al.(2022)Wu, Escontrela, Hafner, Abbeel, and Goldberg]{wu_daydreamer_2022}
Philipp Wu, Alejandro Escontrela, Danijar Hafner, Pieter Abbeel, and Ken Goldberg.
\newblock {DayDreamer}: {World} {Models} for {Physical} {Robot} {Learning}.
\newblock August 2022.
\newblock URL \url{https://openreview.net/forum?id=3RBY8fKjHeu}.

\bibitem[Xu et~al.(2024)Xu, Liu, Sokolsky, Lee, and Kong]{xu_llm-enabled_2024}
Weizhe Xu, Mengyu Liu, Oleg Sokolsky, Insup Lee, and Fanxin Kong.
\newblock {LLM}-enabled {Cyber}-{Physical} {Systems}: {Survey}, {Research} {Opportunities}, and {Challenges}.
\newblock In \emph{International {Workshop} on {Foundation} {Models} for {Cyber}-{Physical} {Systems} \& {Internet} of {Things} ({FMSys})}, May 2024.

\bibitem[Yan et~al.(2024)Yan, Dong, Shao, Lu, Haiyang, Liu, Wang, Wang, Wang, Remondino, and Ma]{yan_renderworld_2024}
Ziyang Yan, Wenzhen Dong, Yihua Shao, Yuhang Lu, Liu Haiyang, Jingwen Liu, Haozhe Wang, Zhe Wang, Yan Wang, Fabio Remondino, and Yuexin Ma.
\newblock {RenderWorld}: {World} {Model} with {Self}-{Supervised} {3D} {Label}, September 2024.
\newblock URL \url{http://arxiv.org/abs/2409.11356}.
\newblock arXiv:2409.11356 [cs].

\bibitem[Yang et~al.(2021)Yang, Liu, Chen, Shen, Hao, and Wang]{yang_causalvae_2021}
Mengyue Yang, Furui Liu, Zhitang Chen, Xinwei Shen, Jianye Hao, and Jun Wang.
\newblock {CausalVAE}: {Disentangled} {Representation} {Learning} via {Neural} {Structural} {Causal} {Models}.
\newblock pages 9593--9602, 2021.
\newblock URL \url{https://openaccess.thecvf.com/content/CVPR2021/html/Yang_CausalVAE_Disentangled_Representation_Learning_via_Neural_Structural_Causal_Models_CVPR_2021_paper.html}.

\bibitem[Yang et~al.(2022)Yang, Zhang, Luu, Ha, Tan, and Yu]{yang_safe_2022}
Tsung-Yen Yang, Tingnan Zhang, Linda Luu, Sehoon Ha, Jie Tan, and Wenhao Yu.
\newblock Safe {Reinforcement} {Learning} for {Legged} {Locomotion}.
\newblock In \emph{2022 {IEEE}/{RSJ} {International} {Conference} on {Intelligent} {Robots} and {Systems} ({IROS})}, pages 2454--2461, October 2022.
\newblock \doi{10.1109/IROS47612.2022.9982038}.
\newblock URL \url{https://ieeexplore.ieee.org/document/9982038}.
\newblock ISSN: 2153-0866.

\bibitem[Zhang et~al.(2020)Zhang, Lyle, Sodhani, Filos, Kwiatkowska, Pineau, Gal, and Precup]{zhang_invariant_2020}
Amy Zhang, Clare Lyle, Shagun Sodhani, Angelos Filos, Marta Kwiatkowska, Joelle Pineau, Yarin Gal, and Doina Precup.
\newblock Invariant {Causal} {Prediction} for {Block} {MDPs}.
\newblock In \emph{Proceedings of the 37th {International} {Conference} on {Machine} {Learning}}, pages 11214--11224. PMLR, November 2020.
\newblock URL \url{https://proceedings.mlr.press/v119/zhang20t.html}.
\newblock ISSN: 2640-3498.

\bibitem[Zhang et~al.(2019)Zhang, Lu, Guo, and Karniadakis]{zhang2019quantifying}
Dongkun Zhang, Lu~Lu, Ling Guo, and George~Em Karniadakis.
\newblock Quantifying total uncertainty in physics-informed neural networks for solving forward and inverse stochastic problems.
\newblock \emph{Journal of Computational Physics}, 397:\penalty0 108850, 2019.

\bibitem[Zhang et~al.(2024)Zhang, Xue, Yan, Zhang, Qiu, Bai, Liu, Cui, and Li]{zhang_efficient_2024}
Haiming Zhang, Ying Xue, Xu~Yan, Jiacheng Zhang, Weichao Qiu, Dongfeng Bai, Bingbing Liu, Shuguang Cui, and Zhen Li.
\newblock An {Efficient} {Occupancy} {World} {Model} via {Decoupled} {Dynamic} {Flow} and {Image}-assisted {Training}, December 2024.
\newblock URL \url{http://arxiv.org/abs/2412.13772}.
\newblock arXiv:2412.13772 [cs].

\bibitem[Zhao et~al.(2024)Zhao, Li, Du, Fu, and Wang]{zhao_physord_2024}
Zhipeng Zhao, Bowen Li, Yi~Du, Taimeng Fu, and Chen Wang.
\newblock {PhysORD}: {A} {Neuro}-{Symbolic} {Approach} for {Physics}-infused {Motion} {Prediction} in {Off}-road {Driving}, October 2024.
\newblock URL \url{http://arxiv.org/abs/2404.01596}.
\newblock arXiv:2404.01596 [cs].

\bibitem[Zheng et~al.(2023)Zheng, Chen, Huang, Zhang, Duan, and Lu]{zheng_occworld_2023}
Wenzhao Zheng, Weiliang Chen, Yuanhui Huang, Borui Zhang, Yueqi Duan, and Jiwen Lu.
\newblock {OccWorld}: {Learning} a {3D} {Occupancy} {World} {Model} for {Autonomous} {Driving}, November 2023.
\newblock URL \url{http://arxiv.org/abs/2311.16038}.
\newblock arXiv:2311.16038 [cs].

\bibitem[Zheng et~al.(2025)Zheng, Li, Ruchkin, Piskac, and Pajic]{zheng_neurostrata_2025}
Xi~Zheng, Ziyang Li, Ivan Ruchkin, Ruzica Piskac, and Miroslav Pajic.
\newblock {NeuroStrata}: {Harnessing} {Neurosymbolic} {Paradigms} for {Improved} {Design}, {Testability}, and {Verifiability} of {Autonomous} {CPS}, February 2025.
\newblock URL \url{http://arxiv.org/abs/2502.12267}.
\newblock arXiv:2502.12267 [cs].

\bibitem[Zhong and Meidani(2023)]{zhong_pi-vae_2023}
Weiheng Zhong and Hadi Meidani.
\newblock {PI}-{VAE}: {Physics}-{Informed} {Variational} {Auto}-{Encoder} for stochastic differential equations.
\newblock \emph{Computer Methods in Applied Mechanics and Engineering}, 403:\penalty0 115664, January 2023.
\newblock ISSN 00457825.
\newblock \doi{10.1016/j.cma.2022.115664}.
\newblock URL \url{https://linkinghub.elsevier.com/retrieve/pii/S0045782522006193}.

\bibitem[Zhuang et~al.(2015)Zhuang, Cheng, Luo, Pan, and He]{zhuang2015supervised}
Fuzhen Zhuang, Xiaohu Cheng, Ping Luo, Sinno~Jialin Pan, and Qing He.
\newblock Supervised representation learning: Transfer learning with deep autoencoders.
\newblock In \emph{Twenty-fourth international joint conference on artificial intelligence}, 2015.

\bibitem[Zuo et~al.(2024)Zuo, Zheng, Huang, Zhou, and Lu]{zuo_gaussianworld_2024}
Sicheng Zuo, Wenzhao Zheng, Yuanhui Huang, Jie Zhou, and Jiwen Lu.
\newblock {GaussianWorld}: {Gaussian} {World} {Model} for {Streaming} {3D} {Occupancy} {Prediction}, December 2024.
\newblock URL \url{http://arxiv.org/abs/2412.10373}.
\newblock arXiv:2412.10373 [cs].

\end{thebibliography}

\clearpage
\appendix
\section*{Appendix}

\subsection*{Experimental Details}

All experiments used an {NVIDIA GeForce RTX 3090 GPU}. The source code can be found at  {\small \url{https://github.com/trustworthy-engineered-autonomy-lab/piwm-principles}}.

\smallskip
\noindent
\textbf{Principles 1--3.} 
Our world model employs a VAE for encoding/decoding high-dimensional image observations and an LSTM time-series predictor for modeling state transitions in the latent space.
The encoder consists of three convolutional layers with increasing feature maps (16, 32, 64) and ReLU activations, downsampling the input image through strided convolutions. The latent representation is parameterized by two fully connected layers (\(\mu\) and \(\log\sigma^2\)), each mapping the encoded feature vector to a latent space of {64 dimensions}. The decoder reconstructs the input image using a fully connected layer followed by three transposed convolutional layers, producing a three-channel output with a sigmoid activation.
The VAE is trained using the {Adam optimizer} with an initial learning rate of {0.001}, incorporating {learning rate decay} to stabilize convergence.

The input to the LSTM consists of {64-dimensional latent representations} extracted by the VAE. The network comprises {two LSTM layers} with a hidden size of {64}, followed by a fully connected output layer mapping to a {64-dimensional output} representing the predicted latent state at the next time step.
The LSTM predictor is trained using the {Adam optimizer} with an initial learning rate of {0.001} and also incorporates {learning rate decay}. The objective is to minimize the MSE between predicted and true latent representations over time.

\noindent
\textbf{Principle 4.} 
Our decoder network maps low-dimensional physical state representations to high-dimensional images using a series of transposed convolutional layers. 
The baseline decoder has one linear layer, two convolutional layers, and a 4-dimensional encoded feature map. Our partitioned decoder only contains one linear layer and one smaller convolutional layer. 
Using a fully connected layer, the decoder first maps the input state (four-dimensional vector in cartpole; eight-dimensional vector in lunar lander) to a high-dimensional feature space. This produces an intermediate representation of size 3×16×24×24. The image output is further refined through independent transposed convolutional layers, each producing a separate image (three independent layers for each segment image for cartpole and lunar lander). The model is trained using the Adam optimizer with an initial learning rate of 0.001.  Training is conducted with mini-batches of size 64, incorporating validation loss tracking to ensure generalization. The loss function is a $\lambda$-weighted combination of the reconstruction MSE of the overall reconstructed image and each segmented part. 

For the partitioned loss function in Definition~\ref{def:world_model_generation}, the choice of $\lambda$ plays a crucial role in image generation behavior:
(a) If $\lambda$ is too small $(< 0.1$), the model fails to separate the three parts, blending ``shadows'' of the original image into the outputs; (b) If $\lambda$ is too big ($> 0.5$), the three parts are completely disconnected, leading to inferior reconstruction quality. 
Through hyperparameter tuning, we found that setting $\lambda = 0.2$ provides an optimal balance between the quality of the separation and the reconstruction in both case studies.

\subsection*{Additional Illustrations}

\begin{figure}[ht]
    \centering
    \includegraphics[width=\textwidth]{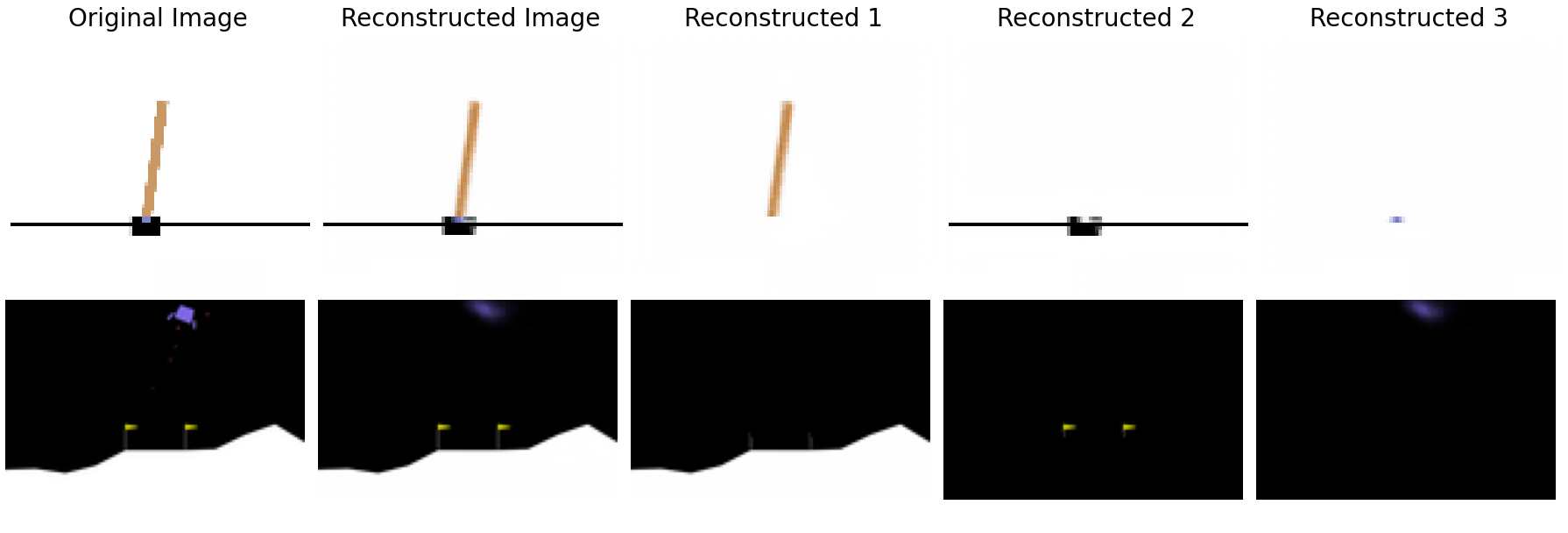}
    \caption{Observations and three reconstructed parts (Principle 4) for the cartpole and lunar lander with $\lambda = 0.2$.}
    \label{fig:parts}
\end{figure}

\begin{figure}[ht] 
    \centering 
    \includegraphics[width=\textwidth]{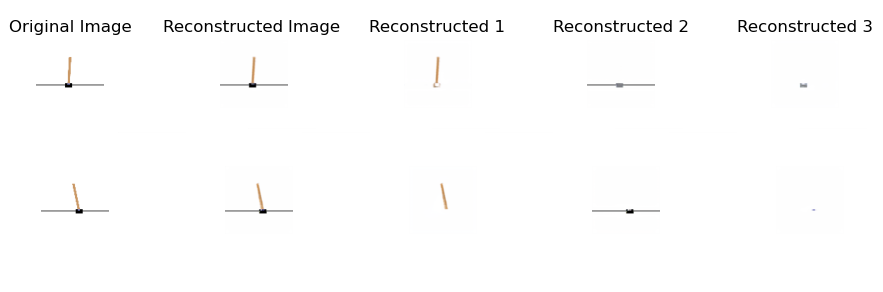} 
    \vspace{-10mm}
    \caption{Imperfect reconstruction for the cart pole: the upper row corresponds to $\lambda=0.01$, while the bottom row corresponds to $\lambda=0.9$.} 
    \label{fig:lambda} 
\end{figure}

\begin{itemize}
    \item Table~\ref{tab:review} lists the literature with the interpretability and adherence to the proposed principles. 
    \item Figure~\ref{fig:parts} shows example observations and their partitioned reconstructions for Principle 4.
    \item Figure~\ref{fig:lambda} shows the imperfect part-wise reconstruction for inadequate values of $\lambda$. 
\end{itemize}

\begin{table}[p]
    \centering
    \begin{adjustbox}{width=\textwidth}
    \begin{tabular}{llcccccc}
       \hline

        \textbf{Short Name} & \textbf{Reference} & \textbf{Principle 1} & \textbf{Principle 2} & \textbf{Principle 3} & \textbf{Principle 4} & \textbf{State interp.} & \textbf{Dyn. interp.} \\
        \hline

        WM & ~\cite[]{ha_recurrent_2018} &  &  &  &  & & \\ 
        
        PlaNet & ~\cite[]{hafner_learning_2019} & \cellcolor{red!20}Weak &  &  &  & &  \cellcolor{red!20}Weak \\  
        
        Dreamer & ~\cite[]{hafner_dream_2020} & \cellcolor{red!20}Weak &  &  &  & & \cellcolor{red!20}Weak \\  
        G-SWM & ~\cite[]{lin_improving_2020} & \cellcolor{green!25}Strong & \cellcolor{red!20}Weak &  & \cellcolor{red!20}Weak &  \cellcolor{yellow!25}Moderate & \cellcolor{red!20}Weak\\  
        AWM & ~\cite[]{kim_active_2020} & \cellcolor{red!20}Weak  &  &  &   & \cellcolor{red!20}Weak & \cellcolor{red!20}Weak \\  %
        Plan2Explore & ~\cite[]{sekar2020planning} & \cellcolor{red!20}Weak &  &  &   & \cellcolor{red!20}Weak & \cellcolor{red!20}Weak \\  

        Pathdreamer & ~\cite[]{koh_pathdreamer_2021} &  &  \cellcolor{red!20}Weak &  \cellcolor{red!20}Weak & \cellcolor{green!25}Strong  & \cellcolor{red!20}Weak & \\  

        DreamerV2 & ~\cite[]{hafner2022mastering} & \cellcolor{red!20}Weak  &  &  &   & \cellcolor{red!20}Weak & \cellcolor{red!20}Weak \\  
        NSV & ~\cite[]{chen_automated_2022} &  & \cellcolor{green!25}Strong &  &   & \cellcolor{green!25}Strong & \\  
        DayDreamer & ~\cite[]{wu_daydreamer_2022} & \cellcolor{red!20}Weak &  &  &   & & \cellcolor{red!20}Weak \\  
        DreamingV2 & ~\cite[]{okada_dreamingv2_2022} & \cellcolor{red!20}Weak &  &  &   & \cellcolor{red!20}Weak & \cellcolor{red!20}Weak \\  
        SEN & ~\cite[]{park_learning_2022} &  & \cellcolor{green!25}Strong  &  &   & \cellcolor{yellow!25}Moderate  &  \\  %
         
        STEDI & ~\cite[]{nakano_interaction-based_2022} & \cellcolor{green!25}Strong & \cellcolor{yellow!25}Moderate &  &   & \cellcolor{green!25}Strong & \\  
        DriveDreamer & ~\cite[]{wang_drivedreamer_2023} &  \cellcolor{red!20}Weak &  & \cellcolor{green!25}Strong &   & \cellcolor{red!20}Weak & \\  
        GAIA-1 & ~\cite[]{hu_gaia-1_2023} & \cellcolor{green!25}Strong &  &  &   &  \cellcolor{yellow!25}Moderate & \\  
        IFactor & ~\cite[]{liu_learning_2023} & \cellcolor{yellow!25}Moderate &  &  &   &  \cellcolor{yellow!25}Moderate & \\  
        IRIS & ~\cite[]{micheli_transformers_2023} & \cellcolor{red!20}Weak &  &  &   & \cellcolor{red!20}Weak & \\  
        MTS3 & ~\cite[]{shaj_multi_nodate} &  \cellcolor{red!20}Weak  &  \cellcolor{red!20}Weak &  &   &  \cellcolor{red!20}Weak & \cellcolor{red!20}Weak\\  
        Denoised MDP & ~\cite[]{wang_denoised_2023} & \cellcolor{yellow!25}Moderate &  &  &   & \cellcolor{yellow!25}Moderate & \\  %
        WM2WM & ~\cite[]{wong_word_2023} &  &  & \cellcolor{green!25}Strong &   &  & \cellcolor{yellow!25}Moderate \\  %
        MWM & ~\cite[]{seo_masked_2023} & \cellcolor{red!20}Weak &  &  &   &  & \cellcolor{red!20}Weak \\  %
        OccWorld & ~\cite[]{zheng_occworld_2023} &  \cellcolor{green!25}Strong & \cellcolor{green!25}Strong &  &   &  \cellcolor{green!25}Strong & \\  
        RAP & ~\cite[]{hao_reasoning_2023} &  &  &  &  &  & \cellcolor{yellow!25}Moderate \\  
        S4WM & ~\cite[]{deng_facing_2023} &  &  &  &   &  & \cellcolor{yellow!25}Moderate \\  
        SWIM & ~\cite[]{mendonca_structured_2023} & \cellcolor{yellow!25}Moderate & \cellcolor{yellow!25}Moderate &  &   &  \cellcolor{yellow!25}Moderate & \\  
        TWM & ~\cite[]{robine_transformer-based_2023} & \cellcolor{red!20}Weak &  &  &  \cellcolor{red!20}Weak &  & \\  
        UniWorld & ~\cite[]{min_uniworld_2023} & & \cellcolor{green!25}Strong & \cellcolor{green!25}Strong &  & \cellcolor{green!25}Strong & \\  
        WorldCloner & ~\cite[]{balloch_neuro-symbolic_2023} &  & \cellcolor{green!25}Strong &  &   &  \cellcolor{yellow!25}Moderate  &  \cellcolor{green!25}Strong \\  
        THICK & ~\cite[]{gumbsch_learning_2023} & \cellcolor{red!20}Weak & \cellcolor{red!20}Weak &  &   &   & \cellcolor{red!20}Weak \\  
         
        AVID & ~\cite[]{rigter_avid_2024} &  &  & \cellcolor{green!25}Strong &  &  & \cellcolor{yellow!25}Moderate \\  
        CMIL & ~\cite[]{kolev_efficient_2024} &  &  & \cellcolor{green!25}Strong &   &   & \cellcolor{red!20}Weak \\  
        DreamerV3 & ~\cite[]{hafner_mastering_2024} & \cellcolor{red!20}Weak & \cellcolor{red!20}Weak &  &   & \cellcolor{red!20}Weak & \cellcolor{red!20}Weak \\  
        DriveWorld & ~\cite[]{min_driveworld_2024} &  & \cellcolor{green!25}Strong  &  &   &  \cellcolor{yellow!25}Moderate &  \cellcolor{red!20}Weak \\  
        DWM & ~\cite[]{ding2024diffusion} &  &  &  &   && \\  
        GaussianWorld & ~\cite[]{zuo_gaussianworld_2024} &  \cellcolor{yellow!25}Moderate & \cellcolor{green!25}Strong &  &   & \cellcolor{green!25}Strong & \cellcolor{yellow!25}Moderate \\  
        Genie & ~\cite[]{bruce_genie_2024} & \cellcolor{yellow!25}Moderate &   &  &  &  \cellcolor{yellow!25}Moderate & \\
        HarmonyWM & ~\cite[]{ma_harmonydream_2024} & \cellcolor{red!20}Weak &  &  &  &  & \cellcolor{red!20}Weak \\  
        OccWM & ~\cite[]{zhang_efficient_2024} & \cellcolor{yellow!25}Moderate & \cellcolor{green!25}Strong &  &   &  \cellcolor{green!25}Strong & \cellcolor{red!20}Weak \\  %
        CovWM & ~\cite[]{popov_mitigating_2024} &  \cellcolor{red!20}Weak & \cellcolor{red!20}Weak &  &   & \cellcolor{red!20}Weak  & \cellcolor{red!20}Weak\\  %
        NWM & ~\cite[]{bar_navigation_2024} &  &  &  &   && \\ 
        OccLLaMA & ~\cite[]{wei_occllama_2024} &  & \cellcolor{green!25}Strong &  &   &  \cellcolor{green!25}Strong & \\  
        PIWM & ~\cite[]{mao_towards_2024} &  & \cellcolor{green!25}Strong & \cellcolor{green!25}Strong  &   & \cellcolor{yellow!25}Moderate & \cellcolor{green!25}Strong  \\  
        R-AIF & ~\cite[]{nguyen_r-aif_2024} &  & \cellcolor{red!20}Weak &  &   &  \cellcolor{red!20}Weak  &  \cellcolor{red!20}Weak \\  
        RenderWorld & ~\cite[]{yan_renderworld_2024} & \cellcolor{green!25}Strong & \cellcolor{green!25}Strong &  &   &  \cellcolor{green!25}Strong & \\  
        Think2Drive & ~\cite[]{li_think2drive_2024} & \cellcolor{red!20}Weak &  &  &   &  &  \cellcolor{red!20}Weak \\  
        TransDreamer & ~\cite[]{chen_transdreamer_2024} & \cellcolor{red!20}Weak &  &  &   &  \cellcolor{red!20}Weak & \cellcolor{red!20}Weak \\  
        VisualPredicator & ~\cite[]{liang_visualpredicator_2024} &  & \cellcolor{green!25}Strong &  &   &  \cellcolor{green!25}Strong & \cellcolor{yellow!25}Moderate \\  
        WorldGPT & ~\cite[]{ge_worldgpt_2024} & \cellcolor{yellow!25}Moderate &  &  & \cellcolor{yellow!25}Moderate  & \cellcolor{yellow!25}Moderate & \cellcolor{yellow!25}Moderate\\  
        Our future vision & & \cellcolor{green!25}Strong & \cellcolor{green!25}Strong & \cellcolor{green!25}Strong & \cellcolor{green!25}Strong  & \cellcolor{green!25}Strong & \cellcolor{green!25}Strong \\  
        
\hline

    \end{tabular}
    \end{adjustbox}
    \caption{Review of notable and state-of-the-art world model architectures for adherence to the \textit{4 principles} and their dynamical/state interpretability.}
    \label{tab:review}
\end{table}




\end{document}